%% file: main.tex
\documentclass[11pt]{article}
\usepackage{acl}  % DO NOT CHANGE THIS
\usepackage{times}
\usepackage{latexsym}

\usepackage[T1]{fontenc}
\usepackage[utf8]{inputenc}

\usepackage{microtype}
\usepackage{algorithm}
\usepackage{algorithmic}
\usepackage{tabularx}
\usepackage{xcolor}
\usepackage{colortbl}
\usepackage{multirow}
\usepackage{arydshln}

\usepackage{xparse}
\usepackage{tikz}
\usetikzlibrary{calc}

\newcommand{\fillcol}{blue!0}
\newcommand{\bordercol}{blue!35}

\newcommand{\setbordercolor}[1]{\renewcommand{\bordercol}{#1}}

\makeatletter
\tikzset{%
     remember picture with id/.style={%
       remember picture,
       overlay,
       save picture id=#1,
     },
     save picture id/.code={%
       \edef\pgf@temp{#1}%
       \immediate\write\pgfutil@auxout{%
         \noexpand\savepointas{\pgf@temp}{\pgfpictureid}}%
     },
     if picture id/.code args={#1#2#3}{%
       \@ifundefined{save@pt@#1}{%
         \pgfkeysalso{#3}%
       }{
         \pgfkeysalso{#2}%
       }
     }
   }

   \def\savepointas#1#2{%
  \expandafter\gdef\csname save@pt@#1\endcsname{#2}%
}

\def\tmk@labeldef#1,#2\@nil{%
  \def\tmk@label{#1}%
  \def\tmk@def{#2}%
}

\tikzdeclarecoordinatesystem{pic}{%
  \pgfutil@in@,{#1}%
  \ifpgfutil@in@%
    \tmk@labeldef#1\@nil
  \else
    \tmk@labeldef#1,(0pt,0pt)\@nil
  \fi
  \@ifundefined{save@pt@\tmk@label}{%
    \tikz@scan@one@point\pgfutil@firstofone\tmk@def
  }{%
  \pgfsys@getposition{\csname save@pt@\tmk@label\endcsname}\save@orig@pic%
  \pgfsys@getposition{\pgfpictureid}\save@this@pic%
  \pgf@process{\pgfpointorigin\save@this@pic}%
  \pgf@xa=\pgf@x
  \pgf@ya=\pgf@y
  \pgf@process{\pgfpointorigin\save@orig@pic}%
  \advance\pgf@x by -\pgf@xa
  \advance\pgf@y by -\pgf@ya
  }%
}

\NewDocumentCommand{\tikzmarkin}{m D(){0.825,-0.10} D(){-0.175,0.27}}{%
      \tikz[remember picture,overlay]
      \draw[line width=1pt,rectangle,fill=\fillcol,draw=\bordercol]
      (pic cs:#1) ++(#2) rectangle (#3)
      ;}

\newcommand\tikzmarkend[2][]{%
\tikz[remember picture with id=#2] #1;}
\usepackage{newfloat}
\usepackage{listings}
\floatstyle{ruled}
\newfloat{listing}{tb}{lst}{}
\floatname{listing}{Listing}
\definecolor{bblue}{HTML}{4F81BD}
\definecolor{rred}{HTML}{C0504D}
\definecolor{ggreen}{HTML}{9BBB59}
\definecolor{ppurple}{HTML}{9F4C7C}
\usepackage{amssymb}

\usepackage{amsmath}
\usepackage{amsthm}
\usepackage{tikz}
\usepackage{pgfplots}
\pgfplotsset{
   compat=1.14,
   legend entry/.initial=,
   every axis plot post/.code={%
       \pgfkeysgetvalue{/pgfplots/legend entry}\tempValue
       \ifx\tempValue\empty
           \pgfkeysalso{/pgfplots/forget plot}%
       \else
           \expandafter\addlegendentry\expandafter{\tempValue}%
       \fi
   },
}

\title{Can Transformers Reason in Fragments of Natural Language?}
\author{Viktor Schlegel $^{1,2}$\\ \texttt{viktor\_schlegel@asus.com}
        \And Kamen V. Pavlov $^{2}$  \\ \texttt{pavloffkamen@gmail.com}
        \AND Ian Pratt-Hartmann $^{2,3}$  \\ \texttt{ian.pratt@manchester.ac.uk}
        \AND $^{1}${\normalfont ASUS Intelligent Cloud Services (AICS), Singapore}\\
        $^{2}${\normalfont Department of Computer Science, University of Manchester, United Kingdom}\\
        $^{3}${\normalfont Instytut Informatyki Uniwersytet Opolski, Poland}
        }

\usepackage{amssymb}
\usepackage{amsmath}
\usepackage{amsthm}
\usepackage{tikz}
\usetikzlibrary{calc,arrows,decorations.pathmorphing,decorations.markings,decorations.pathreplacing}
\input{macros}

\begin{document}

\maketitle

\begin{abstract}
%\marginpar{Small changes}
State-of-the-art 
deep-learning-based approaches to Natural Language Processing (NLP) are credited with various capabilities that involve reasoning with natural language texts. %However, reasoning in this setting is often ill-defined and shallow. 
In this paper we 
%ocus on a setting where reasoning is formally defined. Specifically, we investigate the capability of neural networks to perform satisfiability checks: the capability to %distinguish whether a problem instance---a collection of sentences---represents a logically possible situation. We 
carry out a large-scale empirical study investigating the detection of formally valid inferences in controlled fragments of natural language for which the satisfiability problem becomes increasingly complex. We find that,
while transformer-based language models perform surprisingly well in these scenarios, 
%even on fragments where no efficient symbolic algorithms exist. However, 
a deeper analysis reveals that they appear to overfit to superficial patterns in the data rather than acquiring the logical principles governing the reasoning in these fragments.
%In a detailed analysis, we further find that transformers generate moderately well to systematically longer and more complex problems unseen during training. 
\end{abstract}

\section{Introduction}

%The creative use of reason, like the creative use of language, seems incompatible with the notion that this faculty is rooted in mere habit. For internalizing  a logical principle, like
%mastering an aspect of grammar, requires us to transcend a fixed set of familiar patterns by grasping the possibility of that principle's recursive application. 
%All the more remarkable, therefore, are
The recent success of neural networks in a range of tasks connected with logical inference in natural language is remarkable. Foremost among such systems are those employing transformer-based language models \cite{Vaswani2017}  optimised on large corpora in an unsupervised manner \cite{Devlin2018} and then further fine-tuned on task-specific datasets \cite{Bowman2015,rajpurkar2016squad}.
\ignore{Superficially, the performance of these systems is remarkable.} 
However, concerns  persist regarding so-called ``data-set artefacts''~\cite{gururangan2018annotation,Schlegel2020a,Schlegel2020}. Have netural networks really acquired the principles of reasoning with natural language, or are they merely responding to superficial patterns in the data?

\ignore{Applications of neural networks to inference in formal languages typically employ different architectures---for example graph neural networks encoding logical syntax such as NeuroSAT~\cite{nnFrag:slblmd18} or collections of sampled possible worlds as in PossibleWorldNet~\cite{ nnFrag:esakg18}---or alternatively proceed more obliquely, using neural networks to select inference rules or proof strategies~\cite{nnFrag:kcs17,nnFragrlbs20,nnFrag:ps20,nnFrag:hrwap21}.}

%Popular task settings are Machine Reading Comprehension (MRC) and Natural Language Inference (NLI), with examples of reasoning capabilities ranging from numeric reasoning \cite{Dua2019}, reasoning over entailments in natural language \cite{Bowman2015,Williams2018}, [... couple more examples]. 
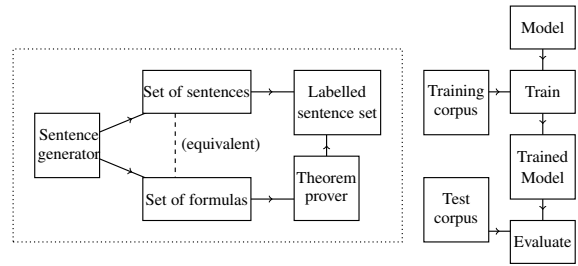
\begin{figure}[!t]
    \centering
    \resizebox{1.02\columnwidth}{!} {
    \input{figure}
    }
    \resizebox{1\columnwidth}{!}{
    \input{figure2}
    }

    \caption{Depiction of our approach to evaluate reasoning capabilities in neural networks.}
    \label{fig:summary}
\end{figure}

In fact, two strands of research may be discerned in recent work on natural language inference (NLI). The first formulates the central task as follows: given a pair of sentences in some natural language, a {\em premise}, $P$ and a {\em hypothesis}, $H$, determine whether 
$P$ (i)  \textit{entails} or (ii) \textit{contradicts} (i.e.~entails the negation of) $H$; or (iii) $P$ is {\em neutral} with respect to $H$. 
For example, the premise
\textit{A person rides his bicycle in the sand beside the ocean} is taken to entail the hypothesis \textit{A person is on a beach}. The primary issue is the possibility of learning such a mapping on the basis of some data set consisting of many such sentence pairs, each  tagged with the a `correct' (i.e.~gold standard) label as determined by human judgement, typically obtained by crowd-sourcing.
% of {\em entails}, {\em contradicts} or {\em neutral}.
Examples include the RTE dataset~\cite{nnFrag:BDDGM09} and the (much larger) SNLI and MNLI datasets~\cite{Bowman2015,Williams2018}. 
%\citet{Bowman2015} reported that 
Neural-network models achieve impressive accuracy on this task \cite{Chen2017EnhancedInference,Devlin2018}. 
%(See also \citet{nnFrag:MMLXZYJ16} and~\citet{nnFrag:RGHKB16} for broadly similar approaches).  A more challenging dataset, MNLI, was presented by~\citet{Williams2018}, with the entailment relation again shown to be learnable using neural letwork models.
%Most strikingly, perhaps, it was shown in~\cite{Devlin2018} that language models computed by bi-directional transformers could be fine-tuned to perform well on the RTE entailment dataset. This latter approach uses a general sequence encoding architecture for the sentences involved, requiring no feature extraction or parsing to generate the vector representations of sentences. 
The nature of the challenge here is (deliberately)  mixed:
on the one hand, solving the NLI problem requires a grasp of the meaning
of various closed-class words  ({\em a}, {\em not}, \dots) 
together with an appreciation of the semantics of the grammatical constructions involved. On the other hand, almost all the entailments or contradictions encountered rely on common sense knowledge (for example, that racing a bicycle involves riding it), which it is the job of the system to acquire. The supposition is that these pieces of commonsense knowledge mediate the entailment of the hypothesis (or its negation)
from the premise. The logical basis of the inferences, so reconstructed, is typically straightforward,
amounting to simple syllogism-like inferences; but the mediating commonsense knowledge is coded in `soft' form, matching the approximate, probabilistic nature of natural language inference.  The operative notion of inference
thus eludes a formal 
definition~\cite{Sugawara2018},% leading to inherent disagreements between the annotators of gold labels \cite{Pavlick2019}, 
ultimately resulting in debatable labels \cite{Schlegel2020}.%\marginpar{Do we really want to say this?}

The%\marginpar{Small changes}  
second strand of research investigates the ability of neural networks to recognize formally valid entailments {\em in sensu stricto}, but nevertheless couched in fragments of natural language. Examples include~\newcite{Salvatore2019,Richardson2019a, nnFrag:rs21} and \newcite{Talmor2020}. Here the focus is on  learning the (potentially complex) inferential patterns inherent in the logical syntax of the closed-class words and grammatical constructions defining the fragment of language under investigation.
For example, the premises
\textit{Every doctor who is not a philosopher is a baker}, {\em John is a doctor} and {\em John is not a baker}
entail the hypothesis \textit{John is a philosopher}.
No commonsense knowledge is required here: the given premises either entail the given hypothesis formally, or they do not. Since entailment in a formally defined fragment of language is a matter of mathematical fact,
not human judgement, inferential problems may be constructed artificially, using random sampling over a grammar, with the correct answers determined by an automated theorem prover (ATP). The question is not whether neural networks can be trained to mimic human annotators' judgements, but rather, whether they can learn the logical principles governing language in question. Note that, because it is formal validity rather than commonsense plausibility that is at issue here, inference tasks of interest typically involve not a \textit{single} premise $P$, 
but rather, a \textit{collection} of premises. 
Individual sentences seldom yield any non-trivial formal entailments.

\ignore{Closer analysis of some widely-used NLI datasets reveals three striking
features of the challenge posed. 
% Crucially, in the NLI challenge, it is the responsibility of the system being evaluated to elicit these meaning postulates---either as a result of its pre-trained language model or fine tuning on the dataset itself. 
Thus, to perform well on this NLI task, a system has to solve two quite different sorts of problem: (i) understanding basic logical syntax, and (ii) inducing meaning postulates. 

The reliance on implicit meaning postulates explains the second striking feature of the NLI

%entailment-{\em pairs} consisting of a {\em single} premise $P$ together a hypothesis $H$. 
Yet, from a logical point of view,  when sentences are short %(as they are in natural language), 
the problem of recognising entailments becomes interesting only when those entailments depend on multiple, interacting premises. Here, of course, %it is 
the induced meaning postulates supply the additional (implicit) premises that render the entailments informative. Yet, %on plausible assumptions about the nature of these implicit premises, 
many of the entailments encountered appear to involve little more than syllogism-like inferences. %corresponding to the medi\ae{}val {\em dictum de omni et nullo}. 
}

These two strands of research reflect a difference in motivation. The former aims to understand inferences apparently performed by humans in everyday linguistic settings---inferences which are necessarily messy and approximate in character. 
From a formal point of view, many of the alleged entailments or contradictions
are in fact judgements of \textit{probability}. Thus, for example the premise {\em Alice is holding a dog} does not, logically speaking, contradict the hypothesis {\em Alice is holding a cat}, even assuming that no dog is a cat; yet such labellings abound in NLI datasets, reflecting the judgements made by (and instructions given to) the human annotators who generate them. 
\ignore{Human annotators have a notoriously weak grasp of the concept of logico-syntactic entailment familiar---for better or for worse---to most logicians, as evidenced by inherent disagreements between different annotators of gold labels \cite{Pavlick2019}, ultimately resulting in data that can be factually wrong \cite{Schlegel2020}. 
Whatever the practical value of teaching a neural network to mimic these judgements, its scientific value must be open to question.}
By contrast, the second (strictly logical) strand of research is motivated principally by a desire to understand the rule-based character of natural language syntax and semantics, and in particular, by the question of whether a neural network can learn the rules in question. 

In this paper, we report on
a large-scale empirical study to investigate whether transformer-based language models can learn the logical syntax of natural language. Specifically, we consider performance on
the problem of determining the satisfiability (logical consistency)
of sets of English sentences featuring
the
determiners
{\em every}, {\em some}, {\em no}, the negative adverb {\em not}, 
and relative clauses,
in the context of a vocabulary of 
count nouns and
 transitive verbs. 
 Importantly, the sentences in question are drawn from various different \textit{fragments} of English, each characterized by a particular range of the grammatical constructions investigated. 
In each case,
data sets---both for training and evaluation---are artificially generated 
in order to test the system's grasp of the underlying logic. Figure~\ref{fig:summary} illustrates the approach, giving an example problem instance for the very simplest of the fragments considered.
We find that
state-of-the-art deep learning-based approaches achieve impressive performance on this task. However, in-depth analysis of their generalisation capabilities reveals a more nuanced pattern, suggesting a tendency to overfit to the parameters that control the data generation, rather than learning the underlying logical principles.

\section{Fragments of Language}

By a {\em fragment} of a natural language, we understand a collection of sentences forming a naturally delineated subset of that language, and equipped with a truth-conditional semantics commanding the assent of its native speakers.
For example, consider the fragment of natural language corresponding to the {\em classical syllogistic} given to us by~\citet[Book A]{nnFrag:aristotlePA}.
In its English-language version, it  consists of the sentence forms
\begin{center}
	\begin{minipage}{10cm}
		\begin{tabbing}
			\small Every $p$ is a $q$ \hspace{2cm} \= 
			\small No $p$ is a $q$ \\
			\small Some $p$ is a $q$               \> 
			\small Some $p$ is not a $q$,
		\end{tabbing}
	\end{minipage}
\end{center}
with schematic variables $p$ and $q$ substituted by common (count) nouns. 
This fragment, which we here denote \sS, can be used to formulate examples such as the one shown in Figure~\ref{fig:summary}.

For present purposes, we may consider the truth conditions of an English sentence to be given by translation to a formal language such as first-order logic, in a way which uncontroversially reconstructs the operative notion of logical entailment. Thus, for example, the sentence forms \sS{}  correspond to the respective logical forms
$\forall x(p(x) \rightarrow \pm q(x))$ and $\exists x(p(x) \wedge \pm q(x))$, where $\pm$ indicates either the presence or absence of negation ($\neg$). An argument in \sS{} is regarded as valid precisely when the first-order translations of its premises entail---in the familiar logical sense---the first-order translation of its conclusion.

Consider now the argument:
\begin{equation}
\text{\begin{minipage}{6cm} \small
\textit{Some artist hates no beekeeper};  \textit{every beekeeper hates some artist}; \textit{therefore some artist is not a beekeeper}.
\end{minipage}}
\label{eq:arg1}
\end{equation}
This argument is again intuitively valid
(though this takes a little thought to see). On the other hand, 
it cannot be sensibly cast in the
syllogistic, because it
so obviously
hinges on intrinsically relational information.
(Observe the alternation of subject and object of the verb {\em hates} in the two premises.) Is there, then, a larger fragment of English in which it might be expressed? Take
the {\em relational syllogistic}, denoted \sR, to be the fragment of English obtained by adding to \sS{} the sentence-forms:
\begin{center}
	\begin{minipage}{10cm}
		\begin{tabbing}
			\small Every $p$ $r$s every/some $q$ \hspace{0.25cm} \= 
			\small Some $p$ $r$s every/some $q$\\
			\small No $p$ $r$s every/any $q$ \> 
			\small Some $p$ does not $r$ every/any $q$,
		\end{tabbing}
	\end{minipage}
\end{center}
where $p$ and $q$ are common count nouns and $r$ a transitive verb. 
As with \sS{}, so too with \sR{}: the truth-conditions of sentences
can be captured by translation to first-order logic in a way which faithfully reconstructs the intuitive notion of validity. 
For example, ``Some artist hates no beekeeper'' may be rendered as $\exists x(\mbox{artist}(x) \wedge \forall y (\mbox{beekeeper}(y) 
\rightarrow \neg \mbox{hate}(x,y)))$, and so on. 
Under these semantics, argument \eqref{eq:arg1} is confirmed as a valid argument in the fragment \sR.

There are other ways to extend the fragment \sS, of course. 
One (very modest) such extension is actually 
featured in the works of~\citet[Ch.~10]{nnFrag:aristotleCdI}. Let us say that the {\em extended classical syllogistic}, denoted \sSd, is the fragment of English which adds to \sS{} the sentence-forms
\begin{center}
	\begin{minipage}{10cm}
		\begin{tabbing}
			\small Every non-$p$ is a $q$ \hspace{1cm} \= 
			\small Some non-$p$ is a $q$\\
			\small No non-$p$ is a $q$                 \> 
			\small Some non-$p$ is not a $q$,
		\end{tabbing}
	\end{minipage}
\end{center}
corresponding to the first-order formulas 
$\forall x(\neg p(x) \rightarrow \pm q(x))$ and $\exists x(\neg p(x) \wedge \pm q(x))$. As we might say, \sSd{} adds `noun-level negation' to \sS. 
Following in the same vein, we can extend \sR{} with noun-level negation, yielding sentences such as 
%\begin{align*}
%& 
\emph{``No non-carpenter admires any non-artist''}
%& & \text{},
%\end{align*}
and so on. We call this fourth fragment the {\em extended relational syllogistic}, denoted \sRd. Again, translation into first-order logic is completely standard.
%Again we ask: what is the computational complexity of determining whether an argument in these fragments of English is valid? 

Let $\cL$ be a fragment of some natural language. A set of \cL-sentences is said to be \textit{satisfiable} if there is a structure making the logical translations of these sentences true. The \textit{satisfiability problem} for $\cL$, denoted $\Sat(\cL)$, is the problem of determining whether a given finite set of sentences of $\cL$ is satisfiable. Provided $\cL$ is equipped with a mechanism for sentence negation (as are all the fragments considered here), any procedure for solving $\Sat(\cL)$ immediately yields a procedure for recognizing logical entailments in $\cL$, since an argument is valid just in case its premises together with the negation of its conclusion is not satisfiable.
Therefore, we have no use for the familiar classification of
natural language inference problems as {\em entailment}, {\em contradiction} and {\em neutral} (and still less the four-way classification of~\citet{nnFrag:tlcxhj21}): the satisfiability problem is as general as we need. As the fragment $\cL$ becomes more expressive, the
corresponding satisfiability problem $\Sat(\cL)$ will, in general, become
more increasingly difficult. However, as we shall see, the details of the resulting trade-off between expressiveness and ease of inference are rather intricate. The focus of the present paper is whether neural networks can learn to solve problems such as $\Sat(\cL)$ for various fragments $\cL$
such as \sS{}, \sSd{}, \sR{} and \sRd{}. 

We remark that the approach taken here is parallel to that taken with respect to the fragments GRL and RCL in~\cite{nnFrag:rs21}.
There are two notable differences, however, Firstly, the fragments considered here are, from a grammatical point of view, more basic, and less clearly a natural language version of propositional logic clauses.
In particular, GRL includes the `sentence' \textit{If carrot and not steak then apples}; and even the more natural RCL is limited to the constructions \textit{Every $X$ who is (not) a $Y$ is (not) a $Z$}.  
To allow comparison between this work and the present study 
we consider the minimal extension of \sS{} by means of \textit{relative clauses}, thus allowing the additional sentence-forms
\begin{center}
	\begin{minipage}{10cm}
		\begin{tabbing}
		\small
			Every $o$ who is a $p$ is a $q$ \hspace{.5cm} \= \small No $o$ who is a $p$ is a $q$  \\
		\small	Some $o$ who is a $p$ is a $q$               \> \small Some $o$ who is a $p$ is not a $q$
		\end{tabbing}
	\end{minipage}
\end{center}
Denote this fragment by \sSrel. Similarly, denote by \sSrelN{} the same fragment additionally allowing negative relative clauses, such as ``Every $o$ who is not a $p$ is a $q$'' and so on. The fragment 
\sSrelN{} is actually an extension of RCL.

An important motivation for the rather more general approach taken here is that 
the satisfiability problem $\Sat(\cL)$ for various fragments $\cL$ of English may be studied from a purely complexity-theoretic point of view. Thus, for example, it is known that the problems $\Sat(\sS)$, $\Sat(\sSd)$ and $\Sat(\sR)$ are \NLogSpace-complete, while $\Sat(\sRd)$ is \ExpTime-complete \cite{nnFrag:p-hm09}. 
Similarly, the satisfiability 
problems for GRL and RCL are easily seen to be \NPTime-complete, as is the problem 
$\Sat(\sSrelN)$; by contrast, the problem 
$\Sat(\sSrel)$
is \PTime-complete~\cite[Theorem 7]{nnFrag:ph14}. 
The question naturally arises as to whether the ability of neural networks to learn to solve these various satisfiability problems correlates with these complexity-theoretic differences. 

\begin{table*}[t]
    \centering
    \begin{tabularx}{1\textwidth}{p{.05\textwidth}XX}
         \textbf{Frag.} & \textbf{Templates} & \textbf{Example sentence} \\
         \hline
         \multirow{2}{*}{\sSd} & Every/No (non-)$p$ is a $q$. & Every artist is a beekeeper. \\
          & Some (non-)$p$ is (not) a $q$. & Some carpenter is not a dentist. \\
         \hdashline
         \multirow{4}{*}{\sR} & \emph{all of \sS{} and:} & \\
          & Every/Some $p$ $r$s every/some $q$. & Every artist chases some beekeeper. \\
          & Some $p$ does not $r$ every/any $q$. & Some beekeeper does not chase any artist. \\
          & No $p$ $r$s every/any $q$. & No beekeeper bewitches any artist. \\
         \hdashline
         \multirow{4}{*}{\sRd} & \emph{all of \sSd{} and:} & \\
          & Every/Some (non-)$p$ $r$s every/some (non-)$q$. & Every non-artist chases some beekeeper. \\
          & Some (non-)$p$ does not $r$ every/any (non-)$q$. & Some beekeeper does not chase any non-artist. \\
          & No (non-)$p$ $r$s every/any (non-) $q$. & No non-beekeeper bewitches any non-artist. \\
          \hdashline
         \multirow{2}{*}{\sSrel} 
          & Every/Some/No $o$ who is a $p$ is a $q$. & Every artist who is a dentist is a carpenter. \\
          & Some $o$ who is a $p$ is not a $q$. & Some dentist who is a hunter is not a spy. \\
          \hdashline
          \multirow{2}{*}{\sSrelN} & \emph{all of \sSrel{} and:} & \\
          & Every/Some/No $o$ who is not a $p$ is a $q$. & Every artist who is a not dentist is a carpenter. \\
          & Some $o$ who is a not $p$ is not a $q$. & Some dentist who is a not hunter is not a spy. \\
          \hline
    \end{tabularx}
    \caption{Templates used to generate the problem instances for all five fragments. Round brackets () denote optionals, forward slashes / denote alternatives.}
    \label{tab:examples}
\end{table*}

Table~\ref{tab:examples}%\marginpar{Main paper? Waddayamean, \textit{upon} acceptance} 
shows all templates used to generate the datasets for each of the fragments. Code to generate the datasets, allowing control of the parameters discussed above, is included in the supplementary material available on github\footnote{\url{https://github.com/schlevik/nlr}}.

\section{Random problems} 
\label{sec:generation}
%\subsection{Problem Generation}
In this section  we outline the construction of collections of sets of formulas in the fragments \sSd, \sR, \sRd, \sSrel{} and \sSrelN, in which each generated set is labelled as \textit{satisfiable} or \textit{unsatisfiable}. (The fragment \sS{} is not interestingly different
from \sSd{}, and will not be considered in the experiments reported here.)%\marginpar{Note added.}
These collections are partitioned into training and evaluation sets, enabling us to test the ability of neural networks to learn to recognise satisfiability under a range of conditions. 

Any sentence in $\sSd$ translates to a formula having either of the forms $\forall x(\pm p(x) \rightarrow \pm q(x))$ or $\exists x(\pm p(x) \wedge \pm q(x))$. We may thus generate a sentence of $\sSd$ pseudo-randomly by selecting a universal sentence with probability $p_u$, a negated subject with probability $p_{\bar s}$ and a negated object with probability $p_{\bar o}$, and then choosing $p$ and $q$ at random from some collection of $n$ nouns. By
carrying out this process $s$ times, we obtain a random set $\Phi$ of \sSd-sentences ($|\Phi| = s$).  For definiteness, we set $p_{\bar s}=p_{\bar o}=0.5$ and $p_u = 0.8$. In addition, we remove any inconsistent sentences, such as ``Some $p$ is not a $p$.''  In 
choosing the various parameters, we fix $n/s=0.8$, which, as we have empirically established,
keeps the proportion of satisfiable instances at roughly 50\%. Collections
of such problem instances are created for various values of $s$.  The ratio $n/s=0.8$ corresponds (for \sSd) roughly to the critical region of SAT problems studied in~\cite{nnFrag:rs21}, 
where it was shown that learning on data from this region is more effective than learning from uniformly sampled data. However, we need to be wary of assuming that such instances are difficult---an issue addressed in Section 4. 

For \sRd, we proceed similarly, generating $s$ sentences at random over a fixed collection of $n$ nouns and $v$ transitive verbs. Every sentence in \sRd{} is either a sentence in \sSd{} or translates to a formula having one of the forms $\forall x (\pm p(x) \rightarrow \gamma)$ or 
$\exists x (\pm p(x) \wedge \gamma)$, where $\gamma$ is either $\forall y(\pm q(y) \rightarrow \pm r(x,y))$ or
$\exists y(\pm q(y) \wedge \pm r(x,y))$. Call sentences in $\sRd{} \setminus  \sSd{}$ {\em relational sentences}. We 
may thus generate a sentence of \sRd{}  pseudo-randomly by choosing to produce a relational sentence with probability 
$p_r$. If we choose to produce a non-relational sentence (i.e.~a sentence in \sSd), we proceed as above; otherwise a negated verb is chosen with probability $p_{\bar v}$, a universally quantified $\gamma$ with probability 
$p_{uu}$; the other parameters, $n$, $v$, $p_u$, $p_s$ and $p_o$ are interpreted as before. By setting  
$p_{\bar s}=p_{\bar o}=0$, we guarantee that every generated relational sentence has a non-negated subject and a non-negated object, and hence is
a sentence of $\sR$.  By repeating this process $s$ times, we obtain an set $\Phi$ of sentences in \sRd{} (or \sR). 
When generating instances of $\Sat(\sR)$, we thus set 
$p_r=0.2$, $p_{\bar s}=p_{\bar o}=0$, 
$p_{\bar v}=0.5$ and $p_u = p_{uu} = 0.8$; in addition, we fix  $n/s=0.6$ and $v/s = 0.15$, which, as we have empirically established,
keeps the ratio of satisfiable to non-satisfiable instances at roughly 50\%. 
Likewise, when generating instances of $\Sat(\sRd)$, we set $p_{\bar s}=p_{\bar v}=0.5$, with the other parameters as for $\sR$; however, we adjust $n/s$ to $0.64$ in order to maintain the ratio of satisfiable to non-satisfiable instances. 

Finally, for fragments with
relative clauses, \sSrel{} and \sSrelN, in addition to the parameters introduced for \sSd, we control the probability of negated relative clauses by the parameter $p_{\bar r}$. By setting  
$p_{\bar r}=0$ we guarantee that every generated sentence has an un-negated relative clause, and hence is
a sentence of \sSrel{}. We additionally set $p_{\bar o}=0.5$ and $p_u=0.8$, and for \sSrelN{}, we set $p_{\bar r}=0.5$.
We empirically establish that setting $(n-0.225)/s~=~0.59$ yields a probability of satisfiability of approximately 0.5
for both fragments.

The satisfiability of a generated problem instance is determined using the Vampire automated theorem prover (ATP) \cite{Kovacs2013First-OrderVampire}, which (assuming termination) either reports that the input set is satisfiable or outputs a proof of a contradiction.
We record, for each generated problem instance, whether it  is satisfiable, and, if not, the number $l$ of lines in the
discovered proof of a contradiction (\emph{proof length}) as well as the number of input sentences, $d$, used in that proof. The ATP terminated on all generated problem instances; there is, however, no general guarantee that the proofs found are the shortest possible.

Generated sentences are created in the relevant fragments of English with templates depicted in Table~\ref{tab:examples}
and realised with dictionaries of nouns that describe categories for unary predicates (e.g. ``artist'', ``beekeeper'') and transitive verbs for binary predicates (e.g. ``admires'', ``bewitches''). We use distinct vocabularies for training and evaluation data and use words with non-overlapping semantic fields.
Unless stated otherwise, to maintain comparability between the different datasets, we generate $60000$ examples for training and $8000$ for evaluating model accuracy. This is achieved by generating $3750$ and $500$ (for training and evaluation sets, respectively) examples for each number of sentences $s$ between $15$ and $30$.

%such as the ratio of relational to syllogistic facts. For a full list of parameters, consult the Appendix. More importantly, we control the probabilities $p_{\bar s}$, $p_{\bar o}$ of negating subjects and objects of relational facts, respectively. 
%With \mbox{$p_{\bar o} = p_{\bar s} = 0$}, we generate problems in $\sR$ and in $\sRd$ otherwise. During generation, we discard problems that are trivially inconsistent, such as $\exists x(\ell(x) \rightarrow \bar{\ell}(x))$. 

\section{Constructed problems} 
One attractive feature of the fragments \sS, \sSd{} and \sR{} is that their satisfiable sets of formulas admit of a simple graph-theoretic characterization. This gives us an additional means of creating data sets comprising challenging problem instances.

We %\marginpar{Small changes} 
illustrate with \sS{}. Let a set $\Phi$ of \sS{}-sentences be given, and let $p_1, \dots, p_n$ be the common nouns (predicates) 
occurring in $\Phi$. Now let $V$ be the set of expressions $p_i(x)$ or $\neg p_i(x)$ ($1 \leq i \leq n$). We call the elments of $V$ {\em literals} and let the variables $\ell$ and $m$ range over $V$. If $\ell \in V$, denote by $\bar{\ell}$ the opposite literal obtained by adding or removing the negation sign as appropriate. Now let $E$ be the set of ordered pairs of literals $(\ell, m)$ such that 
$\Phi$ contains a sentence formalized as
either $\forall x(\ell \rightarrow m)$ or as
$\forall x(\bar{m} \rightarrow \bar{\ell})$.
Thus, $G_\Phi= (V,E)$ is a directed graph. Write
$\ell \Rightarrow_\Phi m$ if there is a path in $G_\Phi$ from $\ell$ to $m$. 
It can be shown 
\citep[Sec.~3]{nnFrag:p-hm09} that a set of \sS-formulas $\Phi$ is satisfiable if and only if $\Phi$ contains no sentence $\exists x (\ell \wedge  m)$ such that either: \textup{(i)} $\ell \Rightarrow_\Phi \bar{\ell}$; \textup{(ii)} $m \Rightarrow_\Phi \bar{m}$ or \textup{(iii)} $\ell \Rightarrow_\Phi \bar{m}$.
Thus, determining (un)satisfiability in \sS{} amounts to detecting certain
{\em forbidden configurations} (in this case: paths) in the directed graph $G_\Phi$. We regard the length of the
path  (if it exists) as the {\em size}
of the forbidden configuration. Satisfiability in \sSd{} is characterized similarly: we simply have to check that, in addition,
$V$ contains no pair of opposite literals 
$o$ and $\bar{o}$ such that $o \Rightarrow \bar{o}$ and $\bar{o} \Rightarrow o$. Again, if a set of sentences of \sSd{} is inconsistent, then it contains a forbidden configuration having a well-defined {\em size}.

This gives us controlled way to generate hard problem instances. Consider the fragment \sSd.
We begin by simply constructing a forbidden configuration having a given size, $d$. To obtain an \textit{unsatisfiable} problem instance $\Phi$ of size $s$, we then add $s - d$ random sentences, checking (using a simple algorithm) that doing so does not create any smaller forbidden configurations. To obtain a \textit{satisfiable} problem instance of size $s$, we reverse one of the implications in the forbidden configuration, and check that the added $s - d$ random sentences do not cause an unsatisfiability. In effect, $d$ is a guaranteed difficulty level; it yields a lower bound on the proof length required to establish unsatisfiability. 
At the same time, the satisfiable and unsatisfiable instances thus generated are not easily distinguished by any superficial characteristics. We denote the problem $\Sat(\sSd)$ constructed as just described with $d$ in a specified range as $\Sat(\sSd_{[\cdot,\cdot]})$. Thus, for example, in $\sSd_{[2,6]}$, unsatisfiable problem instances have $2 \leq d \leq 6$.

For $\sR$,  the corresponding characterization of unsatisfiability in terms of forbidden configurations involves several cases~\citep[Sec.~4]{nnFrag:p-hm09}. 
For simplicity, we generate difficult instances by focusing on just one of these types of forbidden configuration, which we call an $\forall\forall$-configuration.
We begin by constructing an
$\forall\forall$-configuration of specific size $6d$ (for $d \geq 1$).
(Such a collection of sentences is always unsatisfiable.)
To obtain an \textit{unsatisfiable} problem instance of size $s$,
we add $s - 6d$ randomly generated sentences, again checking that 
the resulting unsatisfiability is due entirely to the forbidden configuration.
\textit{Satisfiable} instances are then obtained by reversing one of the implications in the $\forall\forall$-configuration, and checking that this does not lead to unsatisfiability. Denote the problem $\Sat(\sR)$ constructed as just described with $d$ in a specified range as $\Sat(\sR_{\langle\cdot,\cdot\rangle})$. For example, in $\sR_{\langle 1,2\rangle}$, inconsistent problem instances are inconsistent because of $\forall\forall$-configurations of size 6 or 12.%\marginpar{Is this really an or, not a range?}
 
It is not possible to characterize $\Sat(\sRd)$ in terms of forbidden configurations in this simple way. As a substitute, we use the proof-lengths of the proofs found by the ATP as a rough guide to difficulty. 
To generate difficult instances of \sRd, therefore, we first generate random instances as in 
Sec.~\ref{sec:generation};  we then  filter out those unsatisfiable instances with short proof-lengths (as reported by the ATP), and then remove satisfiable instances at random to preserve the proportion of satisfiable instances overall.

\section{Experimental Setup}
%We know from previous studies that neural network models can learn to perform inferences couched in English. 
In the following study, we investigate whether neural networks, and in particular state-of-the-art transformer-based language models \cite{Devlin2018} can learn to perform satisfiability checks on examples representing the selected fragments. First, we investigate whether they can do so in principle, by optimising and evaluating classifiers on training and evaluation data drawn from the same distribution. Second, in an attempt to understand whether they reliably learn the underlying logical principles that govern (un-)satisfiability, we evaluate their \emph{generalisation capabilities} on out-of-distribution evaluation data.  We do so by altering various parameters of the training and evaluation data generation to control the structure %and complexity 
of the generated problem instances.

More specifically, we cast the problem of determining satisfiability as binary text classification and conduct experiments with three transformer based language models, RoBERTa, XLnet and Electra \cite{Liu2019c,Yang2019a,Clark2020ELECTRA:Generators}.which are further described in the Appendix. We employ pre-trained language models because initial experiments showed that these tend to converge faster compared to training from scratch, despite the fact that their pre-training objectives bears little similarity to the task at hand. Following \newcite{Devlin2018}, we represent each problem in plain English text prepended by the special \texttt{[CLS]} token as input to the language model. The text is embedded using the language model, and the embedding of the \texttt{[CLS]} token, as output by the final layer of the language model, is projected into a two-dimensional space, representing the odds of the problem being satisfiable. During inference, we pick the highest logit as the model's prediction and during training we minimise the cross-entropy loss between the logits and the expected class, thus optimising the parameters of the language model to produce the expected prediction conditioned on the input. We keep the choice of hyper-parameters (detailed in Appendix) consistent across experiments.

\section{Results and Analysis}
 % We investigate whether transformer-based language models can learn to perform consistency checks in the discussed fragments and whether they generalise to longer and harder problems. %We inspect the amount of training data required to reach this performance. 
% We furthermore investigate whether this capability generalises to different fragments and, finally, to structurally hard problems by construction. 
%In this section W
We report and analyse the results of the conducted empirical study.
For all results we measure the error as a confidence interval at $\alpha = 0.05$,  using asymptotic normal approximation and omit reportage for brevity, as all measures are in the range of at most
two percent points. We average results obtained for all three language models.
%Comparisons between scores are tested for statistical significance ($p < 0.05$) with Fisher’s exact test.

% \begin{table}[!tb]
%     \centering
%     \begin{tabularx}{0.95\columnwidth}{c|ccccc}
%     \multirow{2}{*}{\textbf{Problem size}} & \multicolumn{5}{c}{\textbf{Fragment}} \\
%     & \sSd & \sR & \sSrel & \sSrelN & \sRd \\
    
%     \hline
%         $15\leq$ \# Sentences $\leq 30$ & 86 & 95 & 96 & 91 & 79 \\
%     \hdashline
%         % Proof length $\geq20$ & & & & \\
%         % Proof length $\geq40$ & & & & \\
%         % \hline
%         % \hline
%         $30\leq$ \# Sentences $\leq 40$\footnotemark & 62 & 92 & 95 & 94 & 76 \\
%         $40\leq$\# Sentences $\leq 45$ & 61 & 90 & - & - & 75 \\
%         \hline
%     \end{tabularx}
%     \caption{Accuracy of \texttt{roberta-large} scaled by 100, trained on the five fragments with problems consisting of $15$ to $30$ sentences and evaluated on longer randomly generated problems in the same fragment.}
%     \label{tab:res-longer}
% \end{table}
\begin{table}[!tb]
    \centering
    \begin{tabular}{c|ccccc}
    % \multicolumn{5}{c}{\textbf{Accuracy}}
    \multirow{1}{*}{\textbf{Problem size}} & \sSd & \sR & \sSrel & \sSrelN & \sRd \\
    
    \hline
        $15\leq s \leq 30$ &  76 & 93 & 94 & 80 & 74\\
    \hdashline
        % Proof length $\geq20$ & & & & \\
        % Proof length $\geq40$ & & & & \\
        % \hline
        % \hline
        $30\leq s \leq 40$\footnotemark[1] & 64 & 89 & 94 & 82 & 72 \\
        $40\leq s \leq 45$ & 63 & 86 & - &  -  & 71 \\
        \hline
    \end{tabular}
    \caption{Accuracy of optimised models trained on random examples consisting of $15$ to $30$ sentences ($s$) and evaluated on longer random problem instances.}
    \label{tab:res-longer}
\end{table}

\footnotetext[1]{$37$ for \sSrel{} and \sSrelN{} to fit transformers' $512$-token limit}
\textbf{\emph{Transformers perform well on random examples in all fragments.}} 
To seek evidence for the first question we train and evaluate separate models on randomly generated problems instances of each of the five fragments. The results are reported the first row of Table~\ref{tab:res-longer} and suggest that the models perform well on randomly generated problems in all fragments, even on the \ExpTime-complete $\sRd{}$ fragment. Surprisingly, the obtained accuracies do not seem to correspond to the complexity classes of the elicited fragments. Note that the results reported in this table are not necessarily that of the best-performing model; for example, by training the RoBERTa model longer and on a larger dataset representing the \sRd{} fragment, we obtain accuracy scores of up to $81\%$ (from original $79\%$, see Table~\ref{tab:res-longer-roberta} in the Appendix). However, to maintain comparability and as the difference is marginal, we use the same training budget for all models optimised on different fragments. The task appears non-trivial, as a simple LSTM-based classifier was not able to outperform the majority class baseline even on the simplest \sSd{} fragment and after explicit hyper-parameter optimisation. 
% For additional experiments on in-fragment generalisation, consult the Appendix.

To investigate whether this performance generalises with the size of the problem instances, we generate evaluation sets with $30 \leq s \leq45$ sentences for longer problems. Note that we are constrained to problems of the size of up to $512$ tokens, as a technical constraint of the pre-trained language model, hence we do not investigate generalisation capabilities to problems beyond 45 (37 for \sSrel{}, \sSrelN{}) sentences.
The remainder of Table~\ref{tab:res-longer} shows that models generalise consistently to problems larger than seen during training, with the notable exception of the model optimised on \sSd{}, which exhibits the most significant drop of over $10$ percent points.

Superficially, Table~\ref{tab:res-longer} appears to indicate good generalisation performance. However, it is important to realise that simply increasing the number of sentences does not make the reasoning problems harder, as witnessed by the fact that the number of sentences, $d$, required to prove contradiction does not increase: on average, $d=3.50$ for examples with  $15$ to $30$ sentences, and  $d=3.58$  for problems with $30$ to $45$ sentences. %While statistically significant, the practical relevance of this difference is negligible. 
Thus, the forbidden configurations the network is identifying are, for the most part, small, even for large numbers of sentences. A complimentary analysis in the Appendix reveals further, that these forbidden configurations are likely to be common between different fragments.

% However, models are capable of picking out these relevant sentences from an increasing number of ``distracting'' sentences not required for the proof.

\textbf{\emph{Transformers are not robust to distribution shifts.}}
Moving on to the second question, we investigate whether the optimised models truly pick up the reasoning patterns as intended or rather overfit to their training data as an artefact of the configuration of parameters controlling the stochastic generation. Approximating the ``hardness'' of a problem instance by its proof length $l$, we find that for all fragments, the overwhelming majority (ranging from 29\% in \sS{} to 86\% in \sSrelN{}) of contradictory examples have short proof lengths of $12$ or less, indicating that random problems are, in fact, unsatisfiable for trivial reasons which are easy to show (See also histogram in Appendix). Thus, we collect ``hard'' examples by \mbox{(over-) generating} a large body of examples and then filtering by proof length and refer to them as \emph{random hard} problem instances. Naturally, in this way we can only capture the hardness of \emph{inconsistent} examples, therefore the evaluation focuses on those.

When comparing the (out-of-distribution) performance of models on ``easy'' and ``hard'' non-satisfiable problem instances with proof length of at least $42$ (Table~\ref{tab:res-harder}, second and third rows), on average, there is a gap of $25$ percent points, %and over $60$ for the model optimised on the \sSrelN{} fragment. 
This suggests that models in fact overfit to simple problems that tend to dominate the
randomly generated datasets without picking up the general principles governing reasoning in the corresponding fragments. Models optimised on ``hard'' training examples  (proof length $\geq 22$, complemented with an equal number of random consistent problems), generalise well to even harder problems (proof length $\geq 42$, Table~\ref{tab:res-harder}, last row).  This suggests that the
models can learn to classify hard problem instances when presented explicitly by supplying appropriate training data. 
This performance does not carry over to simpler problems, however, as models optimised on harder problems exhibit a drop in accuracy when evaluated on simpler problems, as the penultimate row of Table~\ref{tab:res-harder} shows. In conjunction, these observations suggest that the models tend to overfit to patterns in the generated data arising from the parameterisation of the generation algorithm, rather than learning to perform satisfiability checking in a more general sense. In other words,  just like a bad student of logic, they appear to ``learn the proofs'' rather than the logical principles behind the proofs required for successful systematic generalisation. %This is similar to what \newcite{Rozen2019} report about the capability of neural networks to learn the general notion of linguistic phenomena, such as dative alternation. % This analysis suggests 

\begin{table}[!tb]
    \centering
    \resizebox{\columnwidth}{!}{
    \begin{tabular}{cc|ccccc}
    
    \textbf{Train} $l$ & \textbf{Eval} $l$ & \sSd & \sR & \sSrel & \sSrelN & \sRd \\
    
    \hline
         %train $l \geq 6$; eval $l \geq 6$ & 90 & 96 & 98 & 96 & 85 \\
         %\hdashline
         \multirow{3}{*}{\begin{tabular}{@{}c@{}}$l \geq 6$ \\ \emph{(easy)}\end{tabular}} & satisfiable & 61 & 87 & 90 & 53 & 58 \\
          & $l \leq 12$ & 98 & 99 & 99 & 70 & 98\\
         & $l \geq 42$ & 84 & 74 & 70 &  42 & 70 \\
         \hdashline
         \multirow{3}{*}{\begin{tabular}{@{}c@{}}$l \geq 22$ \\ \emph{(hard)}\end{tabular}} &satisfiable & 57 & 89 & 86 & 50 & 52 \\
          & $l \leq 12$ & 68 & 61 &33 & 61& 72\\
         & $l \geq 42$ &100 & 94 & 99 &  96 & 73 \\
        \hline
    \end{tabular}
    }
    \caption{Accuracy of models trained on random satisfiable and easy/hard insatisfiable examples, and evaluated on random satisfiable and easy/hard insatisfiable examples with proof length $l\leq12$ and $l \geq 42$.} %The first row corresponds to the in-fragment performance on inconsistent problems of models reported in Table~\ref{tab:res-random}.}
    \label{tab:res-harder}
\end{table}

\textbf{\emph{Transformers seem unable to reliably learn the distinct reasoning patterns.}} 
Finally, as determining non-satisfiability for the fragments \sSd{} and \sR{} involves detection of one of a handful of forbidden configurations, we generate data that exposes precisely these forbidden configurations. %\marginpar{Better?} 
%This allows us to investigate in more detail, whether the models learn these patterns from random or constructed data and whether previously discussed generalisation behaviour as observed with random hard problems will emerge for constructed examples as well.

The four top rows of Table~\ref{tab:res-constructed} show that models optimised on \emph{random} and \emph{random hard} datasets in \sSd{} and \sR{} pick these patterns up to a varying degree: while all models perform better than chance, the evaluation performance drops considerably, when comparing to in-distribution performance. %The randomly generated \sSd{} data appears to best represent the evaluated reasoning patterns 
In the case of \sSd{}, breaking down the performance by chain depth, as shown in Figure~\ref{fig:breakdown}, reveals that models optimised on \emph{random} data perform best on examples with chain length two (essentially detecting inconsistencies of the form ``All $p$ are $q$'', ``All $q$ are $s$'', ``Some $p$ are not $s$'') and fail to generalise beyond that, while models trained on \emph{random hard} data perform best on examples with chain length five, with their performance deteriorating for examples with longer and shorter chains. This reinforces the previous point that these models appear to have learned to identify problems that have proofs with similar structure to those in their training data rather than learning to solve the problem in a more general sense.

When we both optimise and evaluate models on the constructed datasets, we find that they are capable of learning these inconsistency patterns well, and generalise to harder problems unseen during training: the RoBERTa model optimised on \sSd{} with chain lengths between two and six performs just as well on evaluation data with chain lengths of up to 10 (Table~\ref{tab:res-constructed}, row five). However, other models fail to learn these patterns from constructed \sSd{} data, suggesting that it is a challenging task. Similarly, all models optimised on \sR{} generalise to unseen lengths of the $\forall\forall$-configurations (Table~\ref{tab:res-constructed}, row six). %This suggests that the models can learn to recognise inconsistency patterns when explicitly incentivised by supplying according training data. 
However, even when seemingly learning these patterns, models fail to reliably transfer these capabilities beyond the constructed cases, as evidenced by the poor generalisation performance when trained on the constructed datasets and evaluated on randomly generated data  (Table~\ref{tab:res-constructed}, bottom). For \sR{}, the bad generalisation capability is expected, as  the constructed dataset does not contain inconsistency patterns other than the $\forall\forall$-configurations (e.g. examples where only the non-relational statements are inconsistent). However, this is inconsistent with the case of \sSd{}, as all possible inconsistency patterns are covered in the constructed dataset, yet the optimised RoBERTa models fail to transfer to random \sSd{} data. %Therefore, optimising a model on a combination of constructed \sSd{} and \sR{} examples, improves performance (Table~\ref{tab:res-constructed}, last row). \marginpar{I struggled with this a bit} %Strikingly, One possible explanation is that the constructed \sSd{} examples provide enough signal to detect unsatisfiability patterns in the non-relational part of \sR{} examples, which, as we have shown, dominate the problems in \sR{}. 

\begin{table}[!tb]
    \centering
    \begin{tabular}{l l c}
    \textbf{Train On} & \textbf{Evaluate On} & \textbf{Accuracy} \\
    \hline
    %\multicolumn{3}{c}{\emph{\footnotesize  train random; eval constructed}} \\
    %\hdashline
    \tikzmarkin{d}(1.01,-0.15)(-0.2,0.3)$\sSd$ & \setbordercolor{rred!35}\tikzmarkin{f}(1.2,-0.15)(-0.2,0.3)$\sSd_{[2,6]}$ & $65$ \\ 
    $\sSd_{l\geq22}$ & $\sSd_{[2,6]}$ & $57$ \\
    $\sR$ & $\sR_{\langle1,2\rangle}$ & $53$ \\
    $\sR_{l\geq22}$ \tikzmarkend{d} & $\sR_{\langle1,2\rangle}$ & $58$ \\
    
    %\hline
    %\multicolumn{3}{c}{\emph{\footnotesize train constructed; eval constructed}} \\
    %\hdashline
    \setbordercolor{rred!35}\tikzmarkin{g}(0.1,-0.19)(-0.2,0.3)$\sSd_{[2,6]}$ & $\sSd_{[2,10]}$ &  $96$\footnotemark[2] \\
    %$\sSd_{[2,6]}$ & $\sSd_{[6,10]}$ & $96$ \\
    $\sR_{\langle1,2\rangle}$ & $\sR_{\langle1,3\rangle}$ \tikzmarkend{f} & $100$ \\
    %$\sR_{\langle1,2\rangle}$ & $\sR_{\langle3\rangle}$  & $100$ \\
    %\hline
    %\multicolumn{3}{c}{\emph{\footnotesize  train constructed; eval random}} \\
    %\hdashline
    $\sSd_{[2,6]}$ &\tikzmarkin{e}(1.93,-0.19)(-0.2,0.3)$\sSd$ & $54$\footnotemark[2] \\
    $\sR_{\langle1,2\rangle}$ & $\sR$ & $48$ \\
    %$\sSd_{[2,6]}$+$\sR_{\langle1,2\rangle}$&  $\sSd$& $51$ \\
    $\sSd_{[2,6]}$+$\sR_{\langle1,2\rangle}$\tikzmarkend{g}&$\sR$\tikzmarkend{e}& $88$\footnotemark[2] \\
    
    \hline
%    $\sSd_{[2,6]}$+$\sR_{\langle1,2\rangle}$ &  $\sSd_{[7,10]}$  & $67$ \\
    \end{tabular}
    \caption{Accuracy  of model optimised and evaluated on {\color{bblue} randomly} generated and {\color{rred}constructed} datasets in \sSd and \sR{}. The datasets $\sSd_{l\geq22}$ and $\sR_{l\geq22}$ have unsatisfiable examples with at least proof length $22$, while $\sSd_{[\cdot,\cdot]}$, $\sR_{\langle\cdot,\cdot\rangle}$ have constructed chains in the bracketed range.}
    \label{tab:res-constructed}
\end{table}
\footnotetext[2]{Results for RoBERTa only, as other models failed to converge on constructed \sSd{} data.}

\pgfplotsset{
/pgfplots/my legend/.style={
legend image code/.code={
\draw[thick,ggreen,dotted](-0.05cm,0cm) -- (0.3cm,0cm);%
   }
  }
}

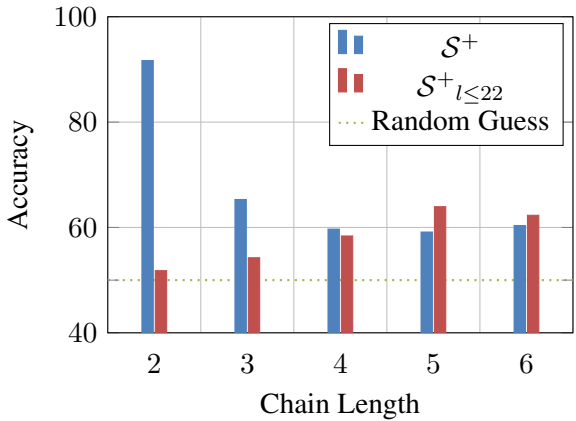
\begin{figure}[tb]
    \centering
    \begin{tikzpicture}
\begin{axis}[
    height=15em,
    width = 1\columnwidth,
    ybar=5pt,
    ymin=40,
    ymax=100,
    scaled y ticks = false,
    extra y ticks = 50,
    extra y tick labels={},
    %y grid style=dashed,
    extra y tick style={grid=major,major grid style={thick,draw=ggreen,dotted}},
    bar width=4.25pt,
    x tick label style={},
    major x tick style = transparent,
    ylabel={Accuracy},
    xlabel={Chain Length},
    x tick label as interval,
    xmajorgrids=true,
    ymajorgrids=true,
    %legend style anchor = north east,
    legend columns=1,
        xticklabels={$2$,$3$,$4$,$5$, $6$},
    xtick={0,1,2,3,4,5},
    xmin=0, xmax=5,
    ]
    
    \addplot+[legend entry=\sSd, bar shift=-2.5pt, color=bblue, fill=bblue] 
    coordinates {
        (0.5,91.69) (1.5,65.31) (2.5,59.69) (3.5,59.13) (4.5,60.37)
        };
    \addplot+[legend entry=\sSd$_{l\leq 22}$, bar shift=+2.5pt, color=rred, fill=rred] 
    coordinates {
        (0.5,51.8) (1.5,54.25) (2.5,58.38) (3.5,63.94) (4.5,62.31)
        };
    \addlegendimage{my legend}
    \addlegendentry{Random Guess}
  %\addlegendentry{Constant Value}
 
\end{axis}
\end{tikzpicture}
    \caption{Accuracy by chain length for RoBERTa models optimised on \sSd{} and \sSd$_{l\leq 22}$, evaluated on \sSd$_{[2,6]}$.
    }
    \label{fig:breakdown}
\end{figure}

%-- train random, eval on constructed: TODO for S, Done for R

%-- train constructed, eval constructed; Done for S, todo for R

% (This is a dump of what we have experiment-wise, before i leave out anything)

% -- train random-gen, eval random-gen: Done for all fragments except S (running) and R+ (running)

% -- train random, eval by depth-filtered: Done for all fragments for S+

% -- train random, eval on longer problems: Done for all fragments

% -- train random filtered, eval random filtered: Done for train min20 and eval min20 min40 for R, R+

% -- train random, eval simpler/harder fragment: Done

%-- train random on subsets: Needs resub for all

%% THUS TODO:
% train-random eval depth-filtered: S (needs sub) R+ (done)
% train random eval longer problems: S (needs sub) R+(done)
% train random eval simpler: R+->R R+->S (submitted)
% train random on subsets: Needs resub for all
% train random eval constructed:  S (awaiting models) R+ (done)
% train constructed, eval constructed: S, R
% ----- 13 exps to submit

% -- What we find when training NNs on our data

% -- What is easy to learn, what is hard to learn?

% -- generalisation dimensions

% -- syllogistic: depth

% -- How does learnability relate to the complexity classes?

\section{Related Work and Conclusion}
%There are several lines of work that are related to our study of NLI.
%We regard the task of NLI from a logical perspective and focus on purely logico-syntactic inferences that do not require the induction of meaning postulates or other general knowledge. We investigate the capability of state-of-the-art neural networks to perform satisfiability checks in controlled fragments of natural language. 

The work reported in this paper represents a continuation of various studies carried out on ``challenge sets'', proposed to investigate the ability of (optimised) neural networks to perform different kinds of reasoning. Such studies range from monotonicity entailments \cite{nnFrag:GRP20,Yanaka2020} to probing lexical knowledge \cite{Glockner2018} to logical connectives \cite{Salvatore2019,Richardson2019a} or verb symmetry \cite{Mitra2019}. These studies are related to ours, in that they seek to isolate capabilities of interest and perform controlled experiments using synthetic datasets. They have in common that models optimised on crowd-sourced datasets, such as MNLI, \cite{Williams2018}, perform poorly on the challenge set data exhibiting the elicited phenomenon, but fine-tuning the optimised model on portions of these data improves the performance. 

%\textbf{Overfitting to superficial patterns:} 
However, as \newcite{Rozen2019} show, good scores after the fine-tuning probably stem from the fact that the investigated model has learned to adapt to the regularities of the challenge set rather than learning a general notion of the investigated phenomenon. Our analysis is similar: transformer-based neural networks perform well on randomly generated data, but this performance is brittle, and the models overfit to the problem space as set out by the dataset generation method. They do not generalise well to examples outside of that space, suggesting that they do not generalise \emph{systematically} \cite{Fodor1988ConnectionismAnalysis}, i.e. they struggle to identify a finite set of rules and to apply them repeatedly. Similar to our findings, research on systematic generalisation suggests that neural networks tend to generalise without systematicity in supervised learning scenarios \cite{Johnson2017CLEVR:Reasoning,Lake2017GeneralizationNetworks,Goodwin2020}, although these studies, unlike ours, did not concern pre-trained models. 

One might argue that it is unfair to expect a statistical model that relies on correlations to learn patterns that are not obviously present in the data. However, it seems that such claims are being made in the literature \cite{Clark2020a}. % For example \newcite postulate transformers capabilities akin to those of a theorem prover, based on the evidence that they can solve previosuly unseen harder problems. 
While we observe that models optimised on constructed examples generalise well to  harder problems unseen during training, we also show that this capability appears not to transfer to examples that are only superficially different. Thus, in our experiments, similar to \newcite{nnFrag:rs21}, we find that the optimised models are not able to reliably disentangle and acquire the different reasoning patterns required to successfully complete the task of determining satisfiability.

Our study highlights one of the issues with empirically postulating neural networks various capabilities by means of good performance on challenge sets. They have only negative predictive power \cite{Gardner2020}: while low performance indicates the lack of a capability, the converse does not necessarily hold. This can be taken as a motivation to develop formal verification methods for neural networks \cite{Shi2020RobustnessTransformers}, or investigate worst-case bounds for different phenomena \cite{Raghunathan2018,Jia2019}.

%We make our code and data publicly available, %\footnote{http://anonymous},
%encouraging other researchers to focus on natural language inference from a logical point of view. 
Our formulated task naturally allows us to expand the scope of the controlled experiments: for example, by increasing the closed-class vocabulary. %increasing the scope of the fragments would offer a way to investigate the understanding of lexical semantics of evaluated systems.\marginpar{Not sure what you mean} 
Another possible avenue is to focus on improving the systematic generalisation of neural approaches, for example by providing the formulas required to prove that a set of sentences is unsatisfiable as additional supervision signal, or by relying on modular or model-based approaches \cite{Andreas2016NeuralNetworks,Lake2013One-shotProcess}.

\section*{Limitations}

The design of the study limits our findings by design - by removing the need of inducing
meaning postulates (i.e. commonsense reasoning and world knowledge) we explicitly focus
on logico-syntactic capabilities, analogous to how “reasoning” is defined for symbolic
approaches to AI. Arguably, In “real world” application scenarios, commonsense reasoning
and world knowledge cannot be fully disconnected from the requirement to perform
reasoning, which allows our inquiry to draw fundamental conclusions about the capabilities
of transformer-based language models rather than to make recommendations which are of
immediate relevance to practitioners.

Our study also suffers from the inductive dilemma. We find that multiple transformer-based
language models follow the trends reported in this paper, specifically that they fail to robustly
identify the reasoning patterns necessary for reliably determining satisfiability in the elicited
fragments. However, due to the empirical nature of this research, this finding is of course not
a guarantee that some neural architecture (transformer-based or otherwise) could still
perform well, when tested in our out-of-distribution evaluation settings.

\section*{Acknowledgements}
The authors would like to acknowledge the use of the Computational Shared Facility at The University of Manchester and thank the anonymous reviewers from the ARR December 2021 cycle for their valuable feedback.

\bibliography{iansContribution}
\bibliographystyle{acl_natbib}

\clearpage
\newpage

\appendix
\section{Additional Details on the Experimental Setup}

Regarding models, We choose RoBERTa, Electra and XLnet as representatives of ``BERTology'' as they all employ different pretraining objectives to obtain the contextualised representations. We do not perform explicit hyperparemeter optimisation for each of the models for each of the datasets, as we are not invested in finding a best performing model, but rather, we are concerned with more general questions about the learnability of the presented problems. We find that our obtained results are similar, and in this regard we expect results to be similar for other transformer-based approaches, as their performance stems from the amount of pre-training data and language model size, rather than architectural choices \cite{Raffel2019}. Finally, investigating multiple models (e.g. more transformer models or hyperparemeter optimisation) increases the amount of computation and thus the carbon footprint \cite{Strubell2019}, which we deem unnecessary given the research questions.

As both data generation and model optimisation are stochastic processes, we are concerned with the impact of chance on the results of our experiments. To investigate whether model convergence is impacted, we optimise five models on \sR{} datasets generated from different random seeds. For model performance, we evaluate one optimised model on five different \sR{} evaluation sets. The results are summarised in Table~\ref{tab:randomness}: the variance of evaluation scores is negligible, while for optimisation, the influence of randomness is more noticeable. The presented loss variance translates to differences in accuracy of up to 2 percent points.
\begin{table}[!b]
    \centering
    \begin{tabular}{l c c c}
    \textbf{Experiment} & \textbf{Metric} & \textbf{Mean} & \textbf{Std. dev.} \\
    \hline
    \emph{train stability} & Loss & $0.07696$ & $0.008$\\
    \emph{eval stability} & Accuracy & $0.953$ & $0.0013$ \\
    \hline
    \end{tabular}
    \caption{Impact of randomness in the data generation process on model optimisation (first row) and on model performance (second row).}
    \label{tab:randomness}
\end{table}
\section{Additional training details}
\begin{figure}[t]
    \centering
    \begin{tikzpicture}
\begin{axis}[
    height=13em,
    width  = 1\columnwidth,
    ybar=5pt,
    ymin=0,
    ymax=80,
    scaled y ticks = false,
    bar width=4.25pt,
    x tick label style={},
    major x tick style = transparent,
    ylabel={Proportion in \%},
    xlabel={proof lengths $d$},
    x tick label as interval,
    xmajorgrids=true,
    ymajorgrids=true,
    y grid style=dashed,
    %legend style anchor = north east,
    legend columns=1,
        xticklabels={$[0;10)$,$[10;20)$,$[20;30)$,$[30;40)$, $[40;\infty)$},
    xtick={0,1,2,3,4,5},
    xmin=0, xmax=5,
    ]
    \addplot+[legend entry=\sSd, bar shift=-10pt, color=pink, fill=pink] 
    coordinates {
        (0.5,18.6) (1.5,63.8) (2.5,15.8) (3.5,1.8) (4.5,0.1)
        };
    \addplot+[legend entry=\sR, bar shift=-5pt, color=rred, fill=rred] 
    coordinates {
        (0.5,24.8) (1.5,67.2) (2.5,6.6) (3.5,1.1) (4.5,0.1)
        };
    \addplot+[legend entry=\sSrel, bar shift=-0pt, color=ggreen, fill=ggreen] 
    coordinates {
        (0.5,60.1) (1.5,35.8) (2.5,3.4) (3.5,0.4) (4.5,0.1)
        };
    \addplot+[legend entry=\sSrelN, bar shift=5pt, color=ppurple, fill=ppurple] 
    coordinates {
        (0.5,79.5) (1.5,16.2) (2.5,2.7) (3.5,0.8) (4.5,0.7)
        };

    \addplot+[legend entry=\sRd, bar shift=10pt, color=bblue, fill=bblue] 
coordinates {
    (0.5,18.3) (1.5,63.6) (2.5,14.2) (3.5,2.6) (4.5,1.1)
    };

\end{axis}
\end{tikzpicture}
    \caption{Proof length distribution.
    }
    \label{fig:hist}
\end{figure}
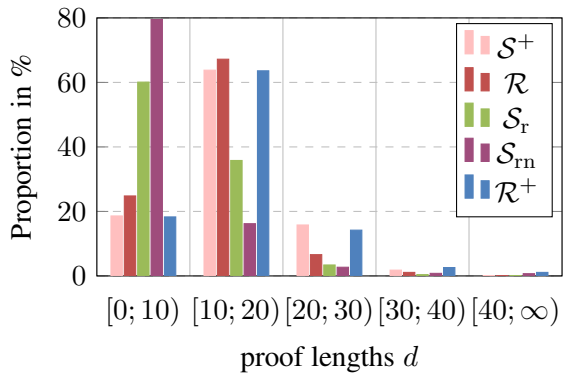
We implement the training and inference in PyTorch 1.10.0 \cite{Paszke2017}. We use the pre-trained language models available in the \texttt{transformers}\footnote{https://github.com/huggingface/transformers} library. We train the \texttt{roberta} model on an Nvidia V100 GPU with 16 GB of RAM. to keep a consistent set of hyperparamters, We train the XLNet models on an Nvidia A100 GPU with 80GB of RAM since these models do not support gradient checkpointing, and the chosen batch size results in large ($\geq$ 16GB) memory requirements during training.

We fix the random seed to maintain deterministic behaviour and the hyper-parameters used for training all models are
\begin{itemize}
    \item \textbf{Batch size:} relying on gradient checkpointing, we are able to set the batch size to $56$. 
    \item \textbf{Learning Rate:} We schedule the learning rate to linearly warm up from zero to $4\cdot10^{-6}$, linearly decaying it to zero, as it was found to perform best across all fragments. We use the ADAM optimiser with the default parameters $\epsilon = 1\times10^{-8}, \beta_1 = 0.99$ and $\beta_2 = 0.999$. Note that this comparatively low learning rate prohibits us from using mixed precision optimisation.
    \item \textbf{Train Epochs:} We train the models for 6 epochs on all fragments to maintain the same training budget.
    \item \textbf{Maximal sequence length}: As the input length varies for different fragments, we ensure that the sequence lengths are set in a way that allows to embed. In practice this varies from $288$ to $432$ tokens. Note that padding input sequences to max length does not impact the training procedure as the padding tokens are not attended to when calculating the embedding of the \verb|[CLS]| input token (nor for any other input tokens).  
\end{itemize}
We find this setting works well for all conducted experiments, thus we keep the same set of hyper-parameters to maintain comparability.
To replicate our experiments, please see the separately supplied code.

\section{Additional Results}
We further investigate whether transformers---perhaps due to their pre-training---can generalise to fragments they have \emph{not} encountered during training. To that end, we evaluate the models on \emph{all} fragments, not only those they have been optimised on. The results are reported in Table~\ref{tab:res-random} and reveal counter-intuitive patterns. Surprisingly, none of the trained models generalises particularly well to the \emph{simpler} \sSd{} fragment (Table~\ref{tab:res-random}, first row). While the results of models trained on the \sSrel{} and \sSrelN{} fragments could be explained by the different problem structure (during training, these models do not encounter sentences as they appear in \sSd{}), the same explanation is not valid for the \sR{} and \sRd{} fragments: problems in these fragments consist of 80\% syllogistic statements on average. In fact, an estimated 90\% of the unsatisfiable problems in \sR{} and \sRd{} contain an inconsistency in the non-relational statements. Therefore, it is reasonable to expect models optimised on \sR{} and \sRd{} to perform well on \sSd, which appears not to be the case. Contrariwise, models optimised on the simpler fragments \sSd{} fragment do generalise well to the harder fragments \sR{} and \sRd{}, as one would expect given their high number of non-relational inconsistencies. This contradictory evidence suggests that the models struggle to reliably identify the configurations leading to insatisfiability. Another observation that eludes a simple explanation is the good performance on the \sR{} fragment of the model optimised on \sSrel, as these fragments are generated from non-overlapping sentence templates. %It was also found (not shown in Table~\ref{tab:res-random}) that optimising a model on a dataset consisting of instances from all fragments does not introduce unexpected synergies: models perform well across the board, without outperforming any of the models optimised on an individual fragment. %suggesting that the model is distracted by the different fragments.
%However, by increasing the training budget, a model can be optimised to perform well on random problems from all fragments.

%namely \sS{} to \sR{} and \sRd{} as well as \sSrel{} to \sSrelN{}, suggests that, for a large fraction of randomly generated problems, a good simple approximating solution might exist, even for harder fragments. 

\begin{table}[!t]
    \resizebox{.99\columnwidth}{!}{%
    \centering
    \begin{tabular}{cc|ccccc}%:c}
    \multicolumn{2}{c|}{\textbf{Evaluated on $\downarrow$}} & \multicolumn{5}{l}{\textbf{Trained on} $\rightarrow$} \\
    & maj. class & \sSd & \sR & \sSrel & \sSrelN & \sRd \\%& all\\
        \hline
        \sSd & $54$ & $\mathbf{86}$ & $61$ & $61$ & $55$ & $70$ \\% & $72$ \\
        \sR  & $52$ &\cellcolor{gray!25} $79$ & $\mathbf{95}$ & $78$  & $54$ & $85$ \\% & 89 \\
        \sSrel & $55$ &\cellcolor{gray!25} $45$ & \cellcolor{gray!25} $48$ & $\mathbf{96}$ & $90$ & $49$ \\%& 93 \\
        \sSrelN & $53$ &\cellcolor{gray!25} $47$  & \cellcolor{gray!25} $50$ & \cellcolor{gray!25} $78$ & $\mathbf{91}$ & $48$ \\% & $94$ \\
        \sRd & $53$ & \cellcolor{gray!25} $73$ & \cellcolor{gray!25} $63$ &  \cellcolor{gray!25} $60$ & \cellcolor{gray!25} $55$ & $\mathbf{79}$ \\% & $74$ \\

    \end{tabular}
    }
    \caption{Accuracy of RoBERTa models evaluated on all fragments. %The diagonal denotes performance on the same fragment and correspond to the first row of Table~\ref{tab:res-random}. 
    Grey results denote generalisation performance to \emph{harder} fragments.} %$\mathcal{B}$ is a majority class baseline.}
    \label{tab:res-random}
   
\end{table}

Training the model on a combined sample of $12000$ problem instances from all fragments results in accuracy scores of $73\%$, $89\%$, $94\%$, $94\%$ and $74\%$, for the fragments \sSd, \sR, \sSrel, \sSrelN{} and \sRd, respectively. 

The distribution of different proof lengths for randomly generated data is shown in Figure~\ref{fig:hist} and supports the hypothesis that most randomly generated inconsistent problem instances are ``easy'' in the sense that they have short proof lengths that lead to refutation.

Figure~\ref{fig:breakdown} shows the breakdown by chain length for models optimised on random and random hard \sSd{} data and evaluated on constructed \sSd{} examples.

Finally, Tables~\ref{tab:res-longer-electra}-\ref{tab:res-constructed-xlnet} report all performances discussed in the main paper broken down by model. We see similar trends across models, with Electra performing best and XLNet often performing worst. We caution to over-interpret these differences, as these could be due to the amount of pre-training of each architecture and the resulting sensitivity to hyper-parameters \cite{Lan2020}.

%ELECTRA TAB 1
\begin{table}[!tb]
    \centering
    \begin{tabular}{c|ccccc}
    % \multicolumn{5}{c}{\textbf{Accuracy}}
    \multirow{1}{*}{\textbf{Problem size}} & \sSd & \sR & \sSrel & \sSrelN & \sRd \\
    
    \hline
        $15\leq s \leq 30$ & 82 & 96 & 95 & 91 & 82  \\
    \hdashline
        $30\leq s \leq 40$ & 72 & 92 & 94 & 93 & 77\\
        $40\leq s \leq 45$ & 70 & 89 & -  &  - &  76 \\
        \hline
    \end{tabular}
    \caption{Accuracy of optimised Electra models trained on random examples consisting of $15$ to $30$ sentences ($s$) and evaluated on longer random problem instances.}
    \label{tab:res-longer-electra}
\end{table}

%ROBERTA TAB1
\begin{table}[!tb]
    \centering
    \begin{tabular}{c|ccccc}
    % \multicolumn{5}{c}{\textbf{Accuracy}}
    \multirow{1}{*}{\textbf{Problem size}} & \sSd & \sR & \sSrel & \sSrelN & \sRd \\
    
    \hline
        $15\leq s \leq 30$ & 86 & 95 & 96 & 91 & 79 \\
    \hdashline
        $30\leq s \leq 40$ & 62 & 92 & 95 & 94 & 76 \\
        $40\leq s \leq 45$ & 61 & 90 & - & - & 75 \\
        \hline
    \end{tabular}
    \caption{Accuracy of optimised RoBERTa models trained on random examples consisting of $15$ to $30$ sentences ($s$) and evaluated on longer random problem instances.}
    \label{tab:res-longer-roberta}
\end{table}

%XLNET TAB 1
\begin{table}[!tb]
    \centering
    \begin{tabular}{c|ccccc}
    % \multicolumn{5}{c}{\textbf{Accuracy}}
    \multirow{1}{*}{\textbf{Problem size}} & \sSd & \sR & \sSrel & \sSrelN & \sRd \\
    
    \hline
        $15\leq s \leq 30$ & 61 & 89 & 92 & 60 & 62  \\
    \hdashline
        $30\leq s \leq 40$ & 60 & 83 & 93 & 60 & 63 \\
        $40\leq s \leq 45$ & 59 & 80 & -  & -  & 64 \\
        \hline
    \end{tabular}
    \caption{Accuracy of optimised XLNet models trained on random examples consisting of $15$ to $30$ sentences ($s$) and evaluated on longer random problem instances.}
    \label{tab:res-longer-xlnet}
\end{table}

% ELECTRA TAB 2
\begin{table}[!tb]
    \centering
    \resizebox{\columnwidth}{!}{
    \begin{tabular}{cc|ccccc}
    
    \textbf{Train} $l$ & \textbf{Eval} $l$ & \sSd & \sR & \sSrel & \sSrelN & \sRd \\
    
    \hline
         %train $l \geq 6$; eval $l \geq 6$ & 90 & 96 & 98 & 96 & 85 \\
         %\hdashline
         \multirow{3}{*}{\begin{tabular}{@{}c@{}}$l \geq 6$ \\ \emph{(easy)}\end{tabular}} & satisfiable & 72 & 92 & 91 & 86 & 71 \\
          & $l \leq 12$ & 99 & 100& 100& 100& 98  \\
         & $l \geq 42$ &  92 & 77 & 79 & 48 & 66 \\
         \hdashline
         \multirow{3}{*}{\begin{tabular}{@{}c@{}}$l \geq 22$ \\ \emph{(hard)}\end{tabular}} &satisfiable &  74 & 92 & 89  & 51 & 65 \\
          & $l \leq 12$ &57 & 64&  30 &  59 & 64\\
         & $l \geq 42$ & 100 & 95 & 99 &  96 & 79 \\
        \hline
    \end{tabular}
    }
    \caption{Accuracy of Electra models trained on random satisfiable and easy/hard insatisfiable examples, and evaluated on random satisfiable and easy/hard insatisfiable examples.} %The first row corresponds to the in-fragment performance on inconsistent problems of models reported in Table~\ref{tab:res-random}.}
    \label{tab:res-harder-electra}
\end{table}

%ROBERTA TAB 2
\begin{table}[!tb]
    \centering
    \resizebox{\columnwidth}{!}{
    \begin{tabular}{cc|ccccc}
    
    \textbf{Train} $l$ & \textbf{Eval} $l$ & \sSd & \sR & \sSrel & \sSrelN & \sRd \\
    
    \hline
         %train $l \geq 6$; eval $l \geq 6$ & 90 & 96 & 98 & 96 & 85 \\
         %\hdashline
         \multirow{3}{*}{\begin{tabular}{@{}c@{}}$l \geq 6$ \\ \emph{(easy)}\end{tabular}} & satisfiable & 78 & 92 & 92 & 87 & 72 \\
          & $l \leq 12$ & 99 & 99 & 99 & 99 & 97 \\
         & $l \geq 42$ & 71 & 68 & 64 & 37 & 56 \\
         \hdashline
         \multirow{3}{*}{\begin{tabular}{@{}c@{}}$l \geq 22$ \\ \emph{(hard)}\end{tabular}} &satisfiable & 60 & 90 & 86 & 48 & 53 \\
          & $l \leq 12$ & 64 & 63 & 28 & 67 & 72\\
         & $l \geq 42$ & 100 & 94 & 98 & 99 & 75 \\
        \hline
    \end{tabular}
    }
    \caption{Accuracy of RoBERTa models trained on random satisfiable and easy/hard insatisfiable examples, and evaluated on random satisfiable and easy/hard insatisfiable examples.} %The first row corresponds to the in-fragment performance on inconsistent problems of models reported in Table~\ref{tab:res-random}.}
    \label{tab:res-harder-roberta}
\end{table}

%XLNET TAB 2
\begin{table}[!tb]
    \centering
    \resizebox{\columnwidth}{!}{
    \begin{tabular}{cc|ccccc}
    
    \textbf{Train} $l$ & \textbf{Eval} $l$ & \sSd & \sR & \sSrel & \sSrelN & \sRd \\
    
    \hline
         %train $l \geq 6$; eval $l \geq 6$ & 90 & 96 & 98 & 96 & 85 \\
         %\hdashline
         \multirow{3}{*}{\begin{tabular}{@{}c@{}}$l \geq 6$ \\ \emph{(easy)}\end{tabular}} & satisfiable & 33 & 77 & 87 & 53 & 31 \\
          & $l \leq 12$ &  95 & 99 & 99 & 70 & 98 \\
         & $l \geq 42$ & 91 & 78 & 68 & 41 & 90 \\
         \hdashline
         \multirow{3}{*}{\begin{tabular}{@{}c@{}}$l \geq 22$ \\ \emph{(hard)}\end{tabular}} &satisfiable & 37 &  85 &  84 &  50 &  38 \\
          & $l \leq 12$ & 84  & 55 & 32 &  57 &  81\\
         & $l \geq 42$ & 100 & 93 & 99  & 92  & 67 \\
        \hline
    \end{tabular}
    }
    \caption{Accuracy of XLNet models trained on random satisfiable and easy/hard insatisfiable examples, and evaluated on random satisfiable and easy/hard insatisfiable examples.} %The first row corresponds to the in-fragment performance on inconsistent problems of models reported in Table~\ref{tab:res-random}.}
    \label{tab:res-harder-xlnet}
\end{table}

%TAB 3 ELECTRA
\begin{table}[!tb]
    \centering
    \begin{tabular}{l l c}
    \textbf{Train On} & \textbf{Evaluate On} & \textbf{Accuracy} \\
    \hline
    %\multicolumn{3}{c}{\emph{\footnotesize  train random; eval constructed}} \\
    %\hdashline
    $\sSd$ & $\sSd_{[2,6]}$ & $70$ \\ 
    $\sSd_{l\geq22}$ & $\sSd_{[2,6]}$ & $58$ \\
    $\sR$ & $\sR_{\langle1,2\rangle}$ & $51$ \\
    $\sR_{l\geq22}$ & $\sR_{\langle1,2\rangle}$ & $61$ \\
    
    %\hline
    %\multicolumn{3}{c}{\emph{\footnotesize train constructed; eval constructed}} \\
    %\hdashline
    $\sSd_{[2,6]}$ & $\sSd_{[2,10]}$ &  $-$ \\
    %$\sSd_{[2,6]}$ & $\sSd_{[6,10]}$ & $96$ \\
    $\sR_{\langle1,2\rangle}$ & $\sR_{\langle1,3\rangle}$ & $100$ \\
    %$\sR_{\langle1,2\rangle}$ & $\sR_{\langle3\rangle}$  & $100$ \\
    %\hline
    %\multicolumn{3}{c}{\emph{\footnotesize  train constructed; eval random}} \\
    %\hdashline
    $\sSd_{[2,6]}$ & $\sSd$ & $-$ \\
    $\sR_{\langle1,2\rangle}$ & $\sR$ & $48$ \\
    %$\sSd_{[2,6]}$+$\sR_{\langle1,2\rangle}$&  $\sSd$& $51$ \\
    $\sSd_{[2,6]}$+$\sR_{\langle1,2\rangle}$&$\sR$& $-$ \\
    
    \hline
%    $\sSd_{[2,6]}$+$\sR_{\langle1,2\rangle}$ &  $\sSd_{[7,10]}$  & $67$ \\
    \end{tabular}
    \caption{Accuracy of Electra models optimised and evaluated on random and constructed datasets in \sSd,\sR{}.}
    \label{tab:res-constructed-electra}
\end{table}

%TAB 3 ROBERTA
\begin{table}[!tb]
    \centering
    \begin{tabular}{l l c}
    \textbf{Train On} & \textbf{Evaluate On} & \textbf{Accuracy} \\
    \hline
    %\multicolumn{3}{c}{\emph{\footnotesize  train random; eval constructed}} \\
    %\hdashline
    $\sSd$ & $\sSd_{[2,6]}$ & $68$ \\ 
    $\sSd_{l\geq22}$ & $\sSd_{[2,6]}$ & $58$ \\
    $\sR$ & $\sR_{\langle1,2\rangle}$ & $57$ \\
    $\sR_{l\geq22}$ & $\sR_{\langle1,2\rangle}$ & $58$ \\
    
    %\hline
    %\multicolumn{3}{c}{\emph{\footnotesize train constructed; eval constructed}} \\
    %\hdashline
    $\sSd_{[2,6]}$ & $\sSd_{[2,10]}$ &  $96$ \\
    %$\sSd_{[2,6]}$ & $\sSd_{[6,10]}$ & $96$ \\
    $\sR_{\langle1,2\rangle}$ & $\sR_{\langle1,3\rangle}$ & $100$ \\
    %$\sR_{\langle1,2\rangle}$ & $\sR_{\langle3\rangle}$  & $100$ \\
    %\hline
    %\multicolumn{3}{c}{\emph{\footnotesize  train constructed; eval random}} \\
    %\hdashline
    $\sSd_{[2,6]}$ & $\sSd$ & $54$ \\
    $\sR_{\langle1,2\rangle}$ & $\sR$ & $48$ \\
    %$\sSd_{[2,6]}$+$\sR_{\langle1,2\rangle}$&  $\sSd$& $51$ \\
    $\sSd_{[2,6]}$+$\sR_{\langle1,2\rangle}$&$\sR$& $88$ \\
    
    \hline
%    $\sSd_{[2,6]}$+$\sR_{\langle1,2\rangle}$ &  $\sSd_{[7,10]}$  & $67$ \\
    \end{tabular}
    \caption{Accuracy of RoBERTa models optimised and evaluated on random and constructed datasets in \sSd,\sR{}.}
    \label{tab:res-constructed-roberta}
\end{table}

%TAB 3 XLNET
\begin{table}[!tb]
    \centering
    \begin{tabular}{l l c}
    \textbf{Train On} & \textbf{Evaluate On} & \textbf{Accuracy} \\
    \hline
    %\multicolumn{3}{c}{\emph{\footnotesize  train random; eval constructed}} \\
    %\hdashline
    $\sSd$ & $\sSd_{[2,6]}$ & $56$ \\ 
    $\sSd_{l\geq22}$ & $\sSd_{[2,6]}$ & $56$ \\
    $\sR$ & $\sR_{\langle1,2\rangle}$ & $52$ \\
    $\sR_{l\geq22}$ & $\sR_{\langle1,2\rangle}$ & $55$ \\
    
    %\hline
    %\multicolumn{3}{c}{\emph{\footnotesize train constructed; eval constructed}} \\
    %\hdashline
    $\sSd_{[2,6]}$ & $\sSd_{[2,10]}$ &  $-$ \\
    %$\sSd_{[2,6]}$ & $\sSd_{[6,10]}$ & $96$ \\
    $\sR_{\langle1,2\rangle}$ & $\sR_{\langle1,3\rangle}$ & $100$ \\
    %$\sR_{\langle1,2\rangle}$ & $\sR_{\langle3\rangle}$  & $100$ \\
    %\hline
    %\multicolumn{3}{c}{\emph{\footnotesize  train constructed; eval random}} \\
    %\hdashline
    $\sSd_{[2,6]}$ & $\sSd$ & $-$ \\
    $\sR_{\langle1,2\rangle}$ & $\sR$ & $48$ \\
    %$\sSd_{[2,6]}$+$\sR_{\langle1,2\rangle}$&  $\sSd$& $51$ \\
    $\sSd_{[2,6]}$+$\sR_{\langle1,2\rangle}$&$\sR$& $-$ \\
    
    \hline
%    $\sSd_{[2,6]}$+$\sR_{\langle1,2\rangle}$ &  $\sSd_{[7,10]}$  & $67$ \\
    \end{tabular}
    \caption{Accuracy of XLNet models optimised and evaluated on random and constructed datasets in \sSd,\sR{}.}
    \label{tab:res-constructed-xlnet}
\end{table}

\section{$\forall\forall$-configurations in \sR{}}
In the fragment \sR{}, satisfiability is characterized by a finite number 
(half a dozen or so) of so-called \textit{forbidden configurations}: families of unsatisfiable sets of formulas, with the instances of each family characterized by a numerical parameter related to that instance's cardinality. It can be shown that any unsatisfiable set of \sR{}-formulas contains an instance of one of these forbidden configurations. The $\forall\forall$-configuration is the most complex of these, and, therefore, the hardest to learn to recognize. Hard unsatisfiable formula sets in \sR{} were constructed in the experiments reported here using the  $\forall\forall$-configuration. We briefly outline its form here. We use abbreviated logical notation, writing 
$\forall(p,q)$ instead of $\forall x(p(x) \rightarrow q(x))$, or
$\forall(p,\exists(q,\neg r))$ instead of $\forall x(p(x) \rightarrow \exists y(\neg r(x,y) \wedge q(y)))$, and so on.

An instance of a $\forall\forall$-configuration with parameter $d$ consists of six sets of $d$ formulas (hence, $6d$ in all). The first two lists entail that all $p$s are $o_1$s and all $p$s are $o_2$s:
\begin{align*}
& \forall(p,p_1), \dots, \forall(p_{d-1},o_1)\\  
& \forall(p,p'_1), \dots, \forall(p'_{d-1},o_2).
\end{align*}
It follows that, if some $p$s exist, then some $o_1$s are $o_2$s. 
The second two lists entail that all $q$s are related by $r$ to
all $o_1$s and to no $o_2$s:
\begin{align*}
& \forall(q,q_1), \dots, 
\forall(q_{d-2},q_{d-1}),
\forall(q_{d-1},\forall(o_1,r))\\  
& \forall(q,q'_1), \dots,
\forall(q'_{d-2},q'_{d-1}),\forall(q'_{d-1},\forall(o_2,\neg r)).
\end{align*}
It follows that, if some $q$s exist, then no $o_1$s are $o_2$s. 
Hence, these four lists entail that, if some $p$s exist, then no $q$s
exist. The final two lists entail that there are both $p$s and $q$s:
\begin{align*}
& \exists(u_0, u_1), \forall(u_1,\exists(u_2,\pm r)), \dots, \forall(u_{d-1},\exists(p, \pm r))\\  
& \exists(u'_0, u'_1), \forall(u'_1,\exists(u'_2, \pm r)), \dots, \forall(u'_{d-1},\exists(q, \pm r)).
\end{align*}
\end{document}

%% file: macros.tex
\newcommand{\ignore}[1]{}

\newcommand{\sS}{\ensuremath{\mathcal{S}}}  % Classical syllogistic ch:syl
\newcommand{\sSd}{\ensuremath{\mathcal{S}^+}}  % extended Cl. syllogistic ch:syl
\newcommand{\sR}{\ensuremath{\mathcal{R}}}  % relational syllogistic ch:syl
\newcommand{\sRd}{\ensuremath{\mathcal{R}^+}}  % extended relational syllogistic ch:syl
\newcommand{\sSrel}{\ensuremath{\mathcal{S}_{\mathrm{r}}}}  % Classical syllogistic with pos rel clauses
\newcommand{\sSrelN}{\ensuremath{\mathcal{S}_{\mathrm{rn}}}}  % Classical syllogistic with all rel clauses

\newcommand{\cL}{\ensuremath{\mathcal{L}}}  % Logic (set of formulas) 

\newcommand{\NLogSpace}{\textsc{NLogSpace}}

\newcommand{\NPTime}{\textsc{NPTime}}

\newcommand{\PTime}{\textsc{PTime}}

\newcommand{\ExpTime}{\textsc{ExpTime}}

\newcommand{\Sat}{\ensuremath{\mbox{\rm Sat}}}

%% file: figure.tex
\tikzset{every picture/.style={line width=0.75pt}} %set default line width to 0.75pt        

\begin{tikzpicture}[x=0.75pt,y=0.75pt,yscale=-1,xscale=1]

%Shape: Rectangle [id:dp9175371653949173] 
\draw   (29.42,30) -- (347.43,30) -- (347.43,90) -- (29.42,90) -- cycle ;
\draw    (29.42,70) -- (347.43,70) ;
\draw    (185,90) -- (185,105) [color={rgb, 255:red, 255; green, 255; blue, 255 }  ][line width=0.75] ;
%Shape: Rectangle [id:dp8710816844993121] 
% \draw   (29.42,110) -- (347.43,110) -- (347.43,130) -- (29.42,130) -- cycle ;
%Shape: Rectangle [id:dp6642257171353615] 
% \draw   (29.14,191) -- (347.15,191) -- (347.15,241) -- (29.14,241) -- cycle ;
% %Straight Lines [id:da9586711568032894] 
% \draw    (185,90) -- (185,108) ;
% \draw [shift={(185,110)}, rotate = 270] [color={rgb, 255:red, 0; green, 0; blue, 0 }  ][line width=0.75]    (10.93,-3.29) .. controls (6.95,-1.4) and (3.31,-0.3) .. (0,0) .. controls (3.31,0.3) and (6.95,1.4) .. (10.93,3.29)   ;
% %Straight Lines [id:da5099259432375362] 
% \draw    (29.42,70) -- (347.43,70) ;
% %Straight Lines [id:da66023425815331] 
% \draw    (184.72,171) -- (184.72,189) ;
% \draw [shift={(184.72,191)}, rotate = 270] [color={rgb, 255:red, 0; green, 0; blue, 0 }  ][line width=0.75]    (10.93,-3.29) .. controls (6.95,-1.4) and (3.31,-0.3) .. (0,0) .. controls (3.31,0.3) and (6.95,1.4) .. (10.93,3.29)   ;
% %Shape: Rectangle [id:dp866530436843754] 
% \draw   (77.95,151) -- (291.48,151) -- (291.48,171) -- (77.95,171) -- cycle ;
% %Straight Lines [id:da2331800152306528] 
% \draw    (184.72,131) -- (184.72,149) ;
% \draw [shift={(184.72,151)}, rotate = 270] [color={rgb, 255:red, 0; green, 0; blue, 0 }  ][line width=0.75]    (10.93,-3.29) .. controls (6.95,-1.4) and (3.31,-0.3) .. (0,0) .. controls (3.31,0.3) and (6.95,1.4) .. (10.93,3.29)   ;

% Text Node
\draw (189.93,22) node   [align=left] {\begin{minipage}[lt]{204.69pt}\setlength\topsep{0pt}
%\begin{center}
{\fontfamily{ptm}\selectfont \textit{{\small Reasoning pattern of interest (e.g. chaining)}}}
%\end{center}

\end{minipage}};
% Text Node
\draw (115.91,50.05) node   [align=left] {\begin{minipage}[lt]{125.4pt}\setlength\topsep{0pt}
{\fontfamily{ptm}\selectfont {\small 1. Every artist is a beekeeper.}}\\{\fontfamily{ptm}\selectfont {\small 2. Every beekeeper is a carpenter.}}
\end{minipage}};
% Text Node
\draw (275.04,49) node   [align=left] {\begin{minipage}[lt]{112.54pt}\setlength\topsep{0pt}
{\fontfamily{ptm}\selectfont {\small 3. No carpenter is a dentist.}}\\{\fontfamily{ptm}\selectfont {\small 4. Some artist is a dentist.}}
\end{minipage}};
% Text Node
% \draw (131.62,102) node  [font=\small] [align=left] {\begin{minipage}[lt]{59.4pt}\setlength\topsep{0pt}
% \begin{flushright}
% {\fontfamily{ptm}\selectfont \textit{Define}}
% \end{flushright}

% \end{minipage}};
% Text Node
% \draw (180.22,121) node   [align=left] {\begin{minipage}[lt]{217.89pt}\setlength\topsep{0pt}
% % \begin{center}
% % {\fontfamily{ptm}\selectfont {\small Specification: Formally defined fragment of Language}}
% % \end{center}

% \end{minipage}};
% Text Node
\draw (184.93,81) node   [align=left] {\begin{minipage}[lt]{211.49pt}\setlength\topsep{0pt}
{\fontfamily{ptm}\selectfont {\small Is this collection of sentences \textit{\textcolor[rgb]{0.96,0.65,0.14}{satisfiable}} or \textcolor[rgb]{0.29,0.56,0.89}{\textit{contradictor}y}?}}
\end{minipage}};
% Text Node
% \draw (189.5,216) node   [align=left] {\begin{minipage}[lt]{218.09pt}\setlength\topsep{0pt}
% {\small {\fontfamily{ptm}\selectfont 1. Can transformers learn to detect contradictory statements?}}\\{\fontfamily{ptm}\selectfont {\small 2. Do they pick up the evaluated general reasoning pattern?}}\\{\fontfamily{ptm}\selectfont {\small 3. If not, can they do so in principle?}}
% \end{minipage}};
% Text Node
% \draw (107.07,183) node  [font=\small] [align=left] {\begin{minipage}[lt]{92.4pt}\setlength\topsep{0pt}
% \begin{flushright}
% {\fontfamily{ptm}\selectfont \textit{Investigate Empirically}}
% \end{flushright}

% \end{minipage}};
% % Text Node
% \draw (184.72,162) node   [align=left] {\begin{minipage}[lt]{145.2pt}\setlength\topsep{0pt}
% \begin{center}
% {\fontfamily{ptm}\selectfont {\small Synthetic dataset (Text Classification)}}
% \end{center}

% \end{minipage}};
% % Text Node
% \draw (131.33,143) node  [font=\small] [align=left] {\begin{minipage}[lt]{59.4pt}\setlength\topsep{0pt}
% \begin{flushright}
% {\fontfamily{ptm}\selectfont \textit{Generate}}
% \end{flushright}

% \end{minipage}};

\end{tikzpicture}

%% file: figure2.tex
\tikzset{every picture/.style={line width=0.75pt}} %set default line width to 0.75pt        

\tikzset{->-/.style={decoration={
markings,
mark=at position #1 with {\arrow{>}}},postaction={decorate}}}
%\begin{center}
\begin{tikzpicture}[scale=0.5]

% Arrows from generator
\draw[->-=0.5] (2.5,6.5) -- (8.5,9);
\draw[->-=0.5] (2.5,6.5) -- (8.5,4);

% Arrows from set of sentences
\draw[->-=0.5] (11,9) -- (13,9);

% Arrows from set of  formulas
\draw[->-=0.5] (11,4) -- (13,4);

% Arrows from Theorem prover
\draw[->-=0.5] (14.5,6) -- (14.5,7);

\draw[dashed] (7.5,8) -- node[right] {(equivalent)} (7.5, 5);

% Generator
\draw[fill= white] (1,5) rectangle (4, 8);
\draw (2.6, 7) node {Sentence};
\draw (2.6, 6) node {generator};

% Set of sentences
\draw[fill= white] (6,8) rectangle (11, 10);
\draw (8.5, 9) node {Set of sentences};

% Set of formulas
\draw[fill= white] (6,3) rectangle (11, 5);
\draw (8.5, 4) node {Set of formulas};

% Theorem prover
\draw[fill= white] (13,3) rectangle (16,6);
\draw (14.5,5) node {Theorem};
\draw (14.5,4) node {prover};

% Labelled sentence set
\draw[fill= white] (13,7) rectangle (17,10);
\draw (15,9) node {Labelled};
\draw (15,8) node {sentence set};

% Big rectangle
\draw[dotted] (0,2)  rectangle (18,11);

% Training corpus
\draw[fill= white] (19,7) rectangle (22,10);
\draw (20.5,9) node {Training};
\draw (20.5,8) node {corpus};

\draw[->-= 0.5] (22,9) -- (23,9);

% Test corpus
\draw[fill= white] (19,2) rectangle (22,5);
\draw (20.5,4) node {Test};
\draw (20.5,3) node {corpus};

\draw[->-= 0.5] (22,2.5) -- (23,2.5);

% NN model
\draw[] (23,11) rectangle (26,13);
\draw (24.5,12) node {Model};

\draw[->-=0.5] (24.5,11) -- (24.5, 10);

% Train
\draw[] (23,8) rectangle (26,10);
\draw (24.5,9) node {Train};

\draw[->-=0.5] (24.5,8) -- (24.5, 7);

% Trained NN model
\draw[] (23,4) rectangle (26,7);
\draw (24.5,6) node {Trained};
\draw (24.5,5) node {Model};

\draw[->-=0.5] (24.5,4) -- (24.5, 3);

% Evaluate
\draw[] (23,1) rectangle (26,3);
\draw (24.5,2) node {Evaluate};

\end{tikzpicture}
%\end{center}

%% file: main.bbl
\begin{thebibliography}{44}
\expandafter\ifx\csname natexlab\endcsname\relax\def\natexlab#1{#1}\fi

\bibitem[{Andreas et~al.(2016)Andreas, Rohrbach, Darrell, and
  Klein}]{Andreas2016NeuralNetworks}
Jacob Andreas, Marcus Rohrbach, Trevor Darrell, and Dan Klein. 2016.
\newblock \href {http://github.com/jacobandreas/nmn2.} {{Neural Module
  Networks}}.

\bibitem[{Aristotle(1963)}]{nnFrag:aristotleCdI}
Aristotle. 1963.
\newblock \emph{Aristotle's Categories and {{\em De Interpretatione}}}.
\newblock Clarendon Press, Oxford.
\newblock (J.R. Ackrill, Tr.).

\bibitem[{Aristotle(1989)}]{nnFrag:aristotlePA}
Aristotle. 1989.
\newblock \emph{Prior Analytics}.
\newblock Hackett, Indianapolis, IN.
\newblock (R. Smith, Tr.).

\bibitem[{Bentivogli et~al.(2009)Bentivogli, Dagan, Dang, Giampiccolo, and
  Magnini}]{nnFrag:BDDGM09}
Luisa Bentivogli, Ido Dagan, Hoa~Trang Dang, Danilo Giampiccolo, and Bernardo
  Magnini. 2009.
\newblock The fifth pascal recognizing textual entailment challenge.
\newblock In \emph{In Proc Text Analysis Conference (TAC’09}.

\bibitem[{Bowman et~al.(2015)Bowman, Angeli, Potts, and Manning}]{Bowman2015}
Samuel~R. Bowman, Gabor Angeli, Christopher Potts, and Christopher~D. Manning.
  2015.
\newblock \href {https://doi.org/10.18653/v1/D15-1075} {{A large annotated
  corpus for learning natural language inference}}.
\newblock In \emph{Proceedings of the 2015 Conference on Empirical Methods in
  Natural Language Processing}, pages 632--642, Stroudsburg, PA, USA.
  Association for Computational Linguistics.

\bibitem[{Chen et~al.(2017)Chen, Zhu, Ling, Wei, Jiang, and
  Inkpen}]{Chen2017EnhancedInference}
Qian Chen, Xiaodan Zhu, Zhen-Hua Ling, Si~Wei, Hui Jiang, and Diana Inkpen.
  2017.
\newblock \href {https://doi.org/10.18653/V1/P17-1152} {{Enhanced LSTM for
  Natural Language Inference}}.
\newblock \emph{ACL 2017 - 55th Annual Meeting of the Association for
  Computational Linguistics, Proceedings of the Conference (Long Papers)},
  1:1657--1668.

\bibitem[{Clark et~al.(2020{\natexlab{a}})Clark, Luong, Brain, Le~Google~Brain,
  and Manning}]{Clark2020ELECTRA:Generators}
Kevin Clark, Minh-Thang Luong, Google Brain, Quoc~V Le~Google~Brain, and
  Christopher~D Manning. 2020{\natexlab{a}}.
\newblock \href {https://doi.org/10.48550/arxiv.2003.10555} {{ELECTRA:
  Pre-training Text Encoders as Discriminators Rather Than Generators}}.
\newblock In \emph{International Conference on Learning Representations
  (ICLR)}.

\bibitem[{Clark et~al.(2020{\natexlab{b}})Clark, Tafjord, and
  Richardson}]{Clark2020a}
Peter Clark, Oyvind Tafjord, and Kyle Richardson. 2020{\natexlab{b}}.
\newblock \href {http://arxiv.org/abs/2002.05867} {{Transformers as Soft
  Reasoners over Language}}.
\newblock In \emph{Proceedings of the Twenty-Ninth International Joint
  Conference on Artificial Intelligence}, pages 3882--3890. International Joint
  Conferences on Artificial Intelligence.

\bibitem[{Devlin et~al.(2019)Devlin, Chang, Lee, and Toutanova}]{Devlin2018}
Jacob Devlin, Ming-Wei Chang, Kenton Lee, and Kristina Toutanova. 2019.
\newblock \href {https://doi.org/10.18653/v1/N19-1423} {{BERT: Pre-training of
  Deep Bidirectional Transformers for Language Understanding}}.
\newblock In \emph{Proceedings of the 2019 Conference of the North American
  Chapter of the Association for Computational Linguistics: Human Language
  Technologies, Volume 1 (Long and Short Papers)}, pages 4171--4186,
  Stroudsburg, PA, USA. Association for Computational Linguistics.

\bibitem[{Fodor and Pylyshyn(1988)}]{Fodor1988ConnectionismAnalysis}
Jerry~A. Fodor and Zenon~W. Pylyshyn. 1988.
\newblock \href {https://doi.org/10.1016/0010-0277(88)90031-5} {{Connectionism
  and cognitive architecture: A critical analysis}}.
\newblock \emph{Cognition}, 28(1-2):3--71.

\bibitem[{Gardner et~al.(2020)Gardner, Artzi, Basmov, Berant, Bogin, Chen,
  Dasigi, Dua, Elazar, Gottumukkala, Gupta, Hajishirzi, Ilharco, Khashabi, Lin,
  Liu, Liu, Mulcaire, Ning, Singh, Smith, Subramanian, Tsarfaty, Wallace,
  Zhang, and Zhou}]{Gardner2020}
Matt Gardner, Yoav Artzi, Victoria Basmov, Jonathan Berant, Ben Bogin, Sihao
  Chen, Pradeep Dasigi, Dheeru Dua, Yanai Elazar, Ananth Gottumukkala, Nitish
  Gupta, Hannaneh Hajishirzi, Gabriel Ilharco, Daniel Khashabi, Kevin Lin,
  Jiangming Liu, Nelson~F. Liu, Phoebe Mulcaire, Qiang Ning, Sameer Singh,
  Noah~A. Smith, Sanjay Subramanian, Reut Tsarfaty, Eric Wallace, Ally Zhang,
  and Ben Zhou. 2020.
\newblock \href {https://doi.org/10.18653/v1/2020.findings-emnlp.117}
  {{Evaluating Models’ Local Decision Boundaries via Contrast Sets}}.
\newblock In \emph{Findings of the Association for Computational Linguistics:
  EMNLP 2020}, pages 1307--1323, Stroudsburg, PA, USA. Association for
  Computational Linguistics.

\bibitem[{Geiger et~al.(2020)Geiger, Richardson, and Potts}]{nnFrag:GRP20}
Atticus Geiger, Kyle Richardson, and Christopher Potts. 2020.
\newblock Neural natural language inference models partially embed theories of
  lexical entailment and negation.
\newblock In \emph{Proceedings of the Third BlackboxNLP Workshop on Analyzing
  and Interpreting Neural Networks for NLP}, pages 163--173. Association for
  Computational Linguistics.

\bibitem[{Glockner et~al.(2018)Glockner, Shwartz, and Goldberg}]{Glockner2018}
Max Glockner, Vered Shwartz, and Yoav Goldberg. 2018.
\newblock \href {https://doi.org/10.18653/v1/P18-2103} {{Breaking NLI Systems
  with Sentences that Require Simple Lexical Inferences}}.
\newblock In \emph{Proceedings of the 56th Annual Meeting of the Association
  for Computational Linguistics (Volume 2: Short Papers)}, pages 650--655,
  Stroudsburg, PA, USA. Association for Computational Linguistics.

\bibitem[{Goodwin et~al.(2020)Goodwin, Sinha, and O’Donnell}]{Goodwin2020}
Emily Goodwin, Koustuv Sinha, and Timothy~J. O’Donnell. 2020.
\newblock \href {https://doi.org/10.18653/v1/2020.acl-main.177} {{Probing
  Linguistic Systematicity}}.
\newblock In \emph{Proceedings of the 58th Annual Meeting of the Association
  for Computational Linguistics}, pages 1958--1969, Stroudsburg, PA, USA.
  Association for Computational Linguistics.

\bibitem[{Gururangan et~al.(2018)Gururangan, Swayamdipta, Levy, Schwartz,
  Bowman, and Smith}]{gururangan2018annotation}
Suchin Gururangan, Swabha Swayamdipta, Omer Levy, Roy Schwartz, Samuel Bowman,
  and Noah~A Smith. 2018.
\newblock \href {https://doi.org/10.18653/v1/N18-2017} {{Annotation Artifacts
  in Natural Language Inference Data}}.
\newblock In \emph{Proceedings of the 2018 Conference of the North American
  Chapter of the Association for Computational Linguistics: Human Language
  Technologies, Volume 2 (Short Papers)}, pages 107--112, Stroudsburg, PA, USA.
  Association for Computational Linguistics.

\bibitem[{Jia et~al.(2019)Jia, Raghunathan, G{\"{o}}ksel, and Liang}]{Jia2019}
Robin Jia, Aditi Raghunathan, Kerem G{\"{o}}ksel, and Percy Liang. 2019.
\newblock \href {https://doi.org/10.18653/v1/D19-1423} {{Certified Robustness
  to Adversarial Word Substitutions}}.
\newblock In \emph{Proceedings of the 2019 Conference on Empirical Methods in
  Natural Language Processing and the 9th International Joint Conference on
  Natural Language Processing (EMNLP-IJCNLP)}, pages 4127--4140, Stroudsburg,
  PA, USA. Association for Computational Linguistics.

\bibitem[{Johnson et~al.(2017)Johnson, Fei-Fei, Hariharan, Zitnick, Van
  Der~Maaten, and Girshick}]{Johnson2017CLEVR:Reasoning}
Justin Johnson, Li~Fei-Fei, Bharath Hariharan, C~Lawrence Zitnick, Laurens Van
  Der~Maaten, and Ross Girshick. 2017.
\newblock {CLEVR: A Diagnostic Dataset for Compositional Language and
  Elementary Visual Reasoning}.
\newblock In \emph{Proceedings of the IEEE Conference on Computer Vision and
  Pattern Recognition (CVPR)}, pages 2901--2910.

\bibitem[{Kov{\'{a}}cs and Voronkov(2013)}]{Kovacs2013First-OrderVampire}
Laura Kov{\'{a}}cs and Andrei Voronkov. 2013.
\newblock \href {https://doi.org/10.1007/978-3-642-39799-8{\_}1} {{First-Order
  Theorem Proving and Vampire}}.
\newblock \emph{Lecture Notes in Computer Science (including subseries Lecture
  Notes in Artificial Intelligence and Lecture Notes in Bioinformatics)}, 8044
  LNCS:1--35.

\bibitem[{Lake and Baroni(2017)}]{Lake2017GeneralizationNetworks}
Brenden~M. Lake and Marco Baroni. 2017.
\newblock \href {http://arxiv.org/abs/1711.00350} {{Generalization without
  systematicity: On the compositional skills of sequence-to-sequence recurrent
  networks}}.
\newblock \emph{35th International Conference on Machine Learning, ICML 2018},
  7:4487--4499.

\bibitem[{Lake et~al.(2013)Lake, Salakhutdinov, and
  Tenenbaum}]{Lake2013One-shotProcess}
Brenden~M. Lake, Ruslan Salakhutdinov, and Joshua~B. Tenenbaum. 2013.
\newblock \href {https://dspace.mit.edu/handle/1721.1/94624} {{One-shot
  learning by inverting a compositional causal process}}.
\newblock \emph{University of Toronoto web domain}, 26.

\bibitem[{Lan et~al.(2020)Lan, Chen, Goodman, Gimpel, Sharma, and
  Soricut}]{Lan2020}
Zhenzhong Lan, Mingda Chen, Sebastian Goodman, Kevin Gimpel, Piyush Sharma, and
  Radu Soricut. 2020.
\newblock \href {http://arxiv.org/abs/1909.11942
  https://openreview.net/forum?id=H1eA7AEtvS} {{ALBERT: A Lite BERT for
  Self-supervised Learning of Language Representations}}.
\newblock In \emph{International Conference on Learning Representations}.

\bibitem[{Liu et~al.(2019)Liu, Ott, Goyal, Du, Joshi, Chen, Levy, Lewis,
  Zettlemoyer, and Stoyanov}]{Liu2019c}
Yinhan Liu, Myle Ott, Naman Goyal, Jingfei Du, Mandar Joshi, Danqi Chen, Omer
  Levy, Mike Lewis, Luke Zettlemoyer, and Veselin Stoyanov. 2019.
\newblock \href {http://arxiv.org/abs/1907.11692} {{RoBERTa: A Robustly
  Optimized BERT Pretraining Approach}}.
\newblock \emph{arXiv preprint arxiv:1907.11692}.

\bibitem[{Mitra et~al.(2020)Mitra, Shrivastava, and Baral}]{Mitra2019}
Arindam Mitra, Ishan Shrivastava, and Chitta Baral. 2020.
\newblock \href {www.aaai.org} {{Enhancing Natural Language Inference Using New
  and Expanded Training Data Sets and New Learning Models}}.
\newblock In \emph{Proceedings of the AAAI Conference on Artificial
  Intelligence}.

\bibitem[{Paszke et~al.(2017)Paszke, Gross, Chintala, Chanan, Yang, DeVito,
  Lin, Desmaison, Antiga, and Lerer}]{Paszke2017}
Adam Paszke, Sam Gross, Soumith Chintala, Gregory Chanan, Edward Yang, Zachary
  DeVito, Zeming Lin, Alban Desmaison, Luca Antiga, and Adam Lerer. 2017.
\newblock \href {https://openreview.net/forum?id=BJJsrmfCZ} {{Automatic
  differentiation in PyTorch}}.
\newblock In \emph{Autodiff Workshop @ NIPS 2017}.

\bibitem[{Pratt-Hartmann and Moss(2009)}]{nnFrag:p-hm09}
I.~Pratt-Hartmann and L.~Moss. 2009.
\newblock Logics for the relational syllogistic.
\newblock \emph{Review of Symbolic Logic}, 2(4):647--683.

\bibitem[{Pratt-Hartmann(2014)}]{nnFrag:ph14}
Ian Pratt-Hartmann. 2014.
\newblock Semantic complexity in natural language.
\newblock In \emph{The Handbook of Contemporary Semantic Theory}, 2nd edition
  edition, pages 429--454. Wiley Blackwell.

\bibitem[{Raffel et~al.(2020)Raffel, Shazeer, Roberts, Lee, Narang, Matena,
  Zhou, Li, and Liu}]{Raffel2019}
Colin Raffel, Noam Shazeer, Adam Roberts, Katherine Lee, Sharan Narang, Michael
  Matena, Yanqi Zhou, Wei Li, and Peter~J. Liu. 2020.
\newblock \href {http://arxiv.org/abs/1910.10683} {{Exploring the Limits of
  Transfer Learning with a Unified Text-to-Text Transformer}}.
\newblock \emph{Journal of Machine Learning Research}, 21:1--67.

\bibitem[{Raghunathan et~al.(2018)Raghunathan, Steinhardt, and
  Liang}]{Raghunathan2018}
Aditi Raghunathan, Jacob Steinhardt, and Percy Liang. 2018.
\newblock {Semidefinite relaxations for certifying robustness to adversarial
  examples}.
\newblock In \emph{Advances in Neural Information Processing Systems 31: Annual
  Conference on Neural Information Processing Systems}, pages 10900--10910.

\bibitem[{Rajpurkar et~al.(2016)Rajpurkar, Zhang, Lopyrev, and
  Liang}]{rajpurkar2016squad}
Pranav Rajpurkar, Jian Zhang, Konstantin Lopyrev, and Percy Liang. 2016.
\newblock \href {https://doi.org/10.18653/v1/D16-1264} {{SQuAD: 100,000+
  Questions for Machine Comprehension of Text}}.
\newblock In \emph{Proceedings of the 2016 Conference on Empirical Methods in
  Natural Language Processing}, pages 2383--2392, Stroudsburg, PA, USA.
  Association for Computational Linguistics.

\bibitem[{Richardson et~al.(2019)Richardson, Hu, Moss, and
  Sabharwal}]{Richardson2019a}
Kyle Richardson, Hai Hu, Lawrence~S. Moss, and Ashish Sabharwal. 2019.
\newblock \href {http://arxiv.org/abs/1909.07521} {{Probing Natural Language
  Inference Models through Semantic Fragments}}.
\newblock In \emph{Proceedings of the AAAI Conference on Artificial
  Intelligence}.

\bibitem[{Richardson and Sabharwal(2021)}]{nnFrag:rs21}
Kyle Richardson and Ashish Sabharwal. 2021.
\newblock \href {https://doi.org/10.48550/ARXIV.2112.09054} {Pushing the limits
  of rule reasoning in transformers through natural language satisfiability}.
\newblock (forthcoming in AAAI 22).

\bibitem[{Rozen et~al.(2019)Rozen, Shwartz, Aharoni, and Dagan}]{Rozen2019}
Ohad Rozen, Vered Shwartz, Roee Aharoni, and Ido Dagan. 2019.
\newblock \href {https://doi.org/10.18653/v1/K19-1019} {{Diversify Your
  Datasets: Analyzing Generalization via Controlled Variance in Adversarial
  Datasets}}.
\newblock In \emph{Proceedings of the 23rd Conference on Computational Natural
  Language Learning (CoNLL)}, pages 196--205, Stroudsburg, PA, USA. Association
  for Computational Linguistics.

\bibitem[{Salvatore et~al.(2019)Salvatore, Finger, and
  Hirata~Jr}]{Salvatore2019}
Felipe Salvatore, Marcelo Finger, and Roberto Hirata~Jr. 2019.
\newblock \href {https://doi.org/10.18653/v1/D19-6103} {{A logical-based corpus
  for cross-lingual evaluation}}.
\newblock In \emph{Proceedings of the 2nd Workshop on Deep Learning Approaches
  for Low-Resource NLP (DeepLo 2019)}, pages 22--30, Stroudsburg, PA, USA.
  Association for Computational Linguistics.

\bibitem[{Schlegel et~al.(2022)Schlegel, Nenadic, and
  Batista-Navarro}]{Schlegel2020a}
Viktor Schlegel, Goran Nenadic, and Riza Batista-Navarro. 2022.
\newblock \href {https://doi.org/10.1017/S1351324922000171} {{A survey of
  methods for revealing and overcoming weaknesses of data-driven Natural
  Language Understanding}}.
\newblock \emph{Natural Language Engineering}, pages 1--31.

\bibitem[{Schlegel et~al.(2020)Schlegel, Valentino, Freitas, Nenadic, and
  Batista-Navarro}]{Schlegel2020}
Viktor Schlegel, Marco Valentino, André~Andre Freitas, Goran Nenadic, and Riza
  Batista-Navarro. 2020.
\newblock \href {http://arxiv.org/abs/2003.04642
  https://www.aclweb.org/anthology/2020.lrec-1.660} {{A Framework for
  Evaluation of Machine Reading Comprehension Gold Standards}}.
\newblock In \emph{Proceedings of The 12th Language Resources and Evaluation
  Conference}, pages 5359--5369, Marseille, France. European Language Resources
  Association.

\bibitem[{Shi et~al.(2020)Shi, Zhang, Chang, Huang, and
  Hsieh}]{Shi2020RobustnessTransformers}
Zhouxing Shi, Huan Zhang, Kai-Wei Chang, Minlie Huang, and Cho-Jui Hsieh. 2020.
\newblock \href {https://doi.org/10.48550/arxiv.2002.06622} {{Robustness
  Verification for Transformers}}.
\newblock In \emph{8th International Conference on Learning Representations,
  ICLR 2018 - Conference Track Proceedings}.

\bibitem[{Strubell et~al.(2019)Strubell, Ganesh, and McCallum}]{Strubell2019}
Emma Strubell, Ananya Ganesh, and Andrew McCallum. 2019.
\newblock \href {https://doi.org/10.18653/v1/P19-1355} {{Energy and Policy
  Considerations for Deep Learning in NLP}}.
\newblock In \emph{Proceedings of the 57th Annual Meeting of the Association
  for Computational Linguistics}, pages 3645--3650, Stroudsburg, PA, USA.
  Association for Computational Linguistics.

\bibitem[{Sugawara et~al.(2018)Sugawara, Inui, Sekine, and
  Aizawa}]{Sugawara2018}
Saku Sugawara, Kentaro Inui, Satoshi Sekine, and Akiko Aizawa. 2018.
\newblock \href {https://doi.org/10.18653/v1/D18-1453} {{What Makes Reading
  Comprehension Questions Easier?}}
\newblock In \emph{Proceedings of the 2018 Conference on Empirical Methods in
  Natural Language Processing}, pages 4208--4219, Stroudsburg, PA, USA.
  Association for Computational Linguistics.

\bibitem[{Talmor et~al.(2020)Talmor, Tafjord, Clark, Goldberg, and
  Berant}]{Talmor2020}
Alon Talmor, Oyvind Tafjord, Peter Clark, Yoav Goldberg, and Jonathan Berant.
  2020.
\newblock \href {http://arxiv.org/abs/2006.06609} {{Leap-Of-Thought: Teaching
  Pre-Trained Models to Systematically Reason Over Implicit Knowledge}}.
\newblock In \emph{Advances in Neural Information Processing Systems 33: Annual
  Conference on Neural Information Processing Systems}.

\bibitem[{Tian et~al.(2021)Tian, Li, Chen, Xiao, He, and Jin}]{nnFrag:tlcxhj21}
Jidong Tian, Yitian Li, Wenqing Chen, Liqiang Xiao, Hao He, and Yaohui Jin.
  2021.
\newblock \href {https://aclanthology.org/2021.emnlp-main.303} {Diagnosing the
  first-order logical reasoning ability through {L}ogic{NLI}}.
\newblock In \emph{Proceedings of the 2021 Conference on Empirical Methods in
  Natural Language Processing}, pages 3738--3747, Online and Punta Cana,
  Dominican Republic. Association for Computational Linguistics.

\bibitem[{Vaswani et~al.(2017)Vaswani, Shazeer, Parmar, Uszkoreit, Jones,
  Gomez, Kaiser, and Polosukhin}]{Vaswani2017}
Ashish Vaswani, Noam Shazeer, Niki Parmar, Jakob Uszkoreit, Llion Jones,
  Aidan~N. Gomez, Lukasz Kaiser, and Illia Polosukhin. 2017.
\newblock \href {http://arxiv.org/abs/1706.03762
  http://papers.nips.cc/paper/7181-attention-is-all-you-need.pdf} {{Attention
  Is All You Need}}.
\newblock In \emph{Advances in Neural Information Processing Systems 30}, pages
  5998--6008.

\bibitem[{Williams et~al.(2018)Williams, Nangia, and Bowman}]{Williams2018}
Adina Williams, Nikita Nangia, and Samuel Bowman. 2018.
\newblock \href {https://doi.org/10.18653/v1/N18-1101} {{A Broad-Coverage
  Challenge Corpus for Sentence Understanding through Inference}}.
\newblock In \emph{Proceedings of the 2018 Conference of the North American
  Chapter of the Association for Computational Linguistics: Human Language
  Technologies, Volume 1 (Long Papers)}, pages 1112--1122, Stroudsburg, PA,
  USA. Association for Computational Linguistics.

\bibitem[{Yanaka et~al.(2020)Yanaka, Mineshima, Bekki, and Inui}]{Yanaka2020}
Hitomi Yanaka, Koji Mineshima, Daisuke Bekki, and Kentaro Inui. 2020.
\newblock \href {https://doi.org/10.18653/v1/2020.acl-main.543} {{Do Neural
  Models Learn Systematicity of Monotonicity Inference in Natural Language?}}
\newblock In \emph{Proceedings of the 58th Annual Meeting of the Association
  for Computational Linguistics}, pages 6105--6117, Stroudsburg, PA, USA.
  Association for Computational Linguistics.

\bibitem[{Yang et~al.(2019)Yang, Dai, Yang, Carbonell, Salakhutdinov, and
  Le}]{Yang2019a}
Zhilin Yang, Zihang Dai, Yiming Yang, Jaime Carbonell, Ruslan Salakhutdinov,
  and Quoc~V Le. 2019.
\newblock \href {https://github.com/zihangdai/xlnet} {{XLNet: Generalized
  Autoregressive Pretraining for Language Understanding}}.
\newblock In \emph{In Proceedings of Advances in Neural Information Processing
  Systems 32 (NeurIPS 2019)}, pages 5753--5763.

\end{thebibliography}
